\documentclass[10pt,english,pre,nofootinbib,reprint,showpacs,showkeys,preprintnumbers,floatfix]{revtex4-1}
\usepackage[latin9]{inputenc}
\usepackage[a4paper]{geometry}
\usepackage{xcolor}
\usepackage{pdfcolmk}
\usepackage{babel}
\usepackage{units}
\usepackage{textcomp}
\usepackage{mathrsfs}
\usepackage{amsmath}
\usepackage{amssymb}
\usepackage{stmaryrd}
\usepackage{graphicx}
\usepackage{esint}
\PassOptionsToPackage{normalem}{ulem}
\usepackage{ulem}
\usepackage[unicode=true,pdfusetitle,
 bookmarks=true,bookmarksnumbered=false,bookmarksopen=false,
 breaklinks=false,pdfborder={0 0 1},backref=false,colorlinks=false]
 {hyperref}

\makeatletter

\providecolor{lyxadded}{rgb}{0,0,1}
\providecolor{lyxdeleted}{rgb}{1,0,0}
\DeclareRobustCommand{\lyxdeleted}[3]{{\texorpdfstring{\color{lyxdeleted}\sout{#3}}{}}}

 \@ifundefined{textcolor}{}
 {%
   \definecolor{BLACK}{gray}{0}
   \definecolor{WHITE}{gray}{1}
   \definecolor{RED}{rgb}{1,0,0}
   \definecolor{GREEN}{rgb}{0,1,0}
   \definecolor{BLUE}{rgb}{0,0,1}
   \definecolor{CYAN}{cmyk}{1,0,0,0}
   \definecolor{MAGENTA}{cmyk}{0,1,0,0}
   \definecolor{YELLOW}{cmyk}{0,0,1,0}
 }

\usepackage{aas_macros}
\usepackage{bbold}

\renewcommand{\lyxdeleted}[3]{}

\let\textquotedbl="

\makeatother

\begin{document}

\title{Correlated signal inference by free energy exploration}

\author{Torsten A. Enßlin, Jakob Knollmüller}

\affiliation{{\small{}Max-Planck-Institut für Astrophysik, Karl-Schwarzschildstr.~1,
85748 Garching, Germany}\\
Ludwig-Maximilians-Universität München, Geschwister-Scholl-Platz{\small{}~}1,
80539 Munich, Germany}
\begin{abstract}
The inference of correlated signal fields with unknown correlation
structures is of high scientific and technological relevance, but
poses significant conceptual and numerical challenges. To address
these, we develop the \textit{correlated signal inference }(CSI) algorithm
within \textit{information field theory} (IFT) and discuss its numerical
implementation. To this end, we introduce the \textit{free energy
exploration} (\textsc{FrEE}) strategy for \textit{numerical information
field theory} (\textsc{NIFTy}) applications. The \textsc{FrEE} strategy
is to let the mathematical structure of the inference problem determine
the dynamics of the numerical solver. \textsc{FrEE} uses the Gibbs
free energy formalism for all involved unknown fields and correlation
structures without marginalization of nuisance quantities. It thereby
avoids the complexity marginalization often impose to IFT equations.
\textsc{FrEE} simultaneously solves for the mean and the uncertainties
of signal, nuisance, and auxiliary fields, while exploiting any analytically
calculable quantity.  Here, we develop the \textsc{FrEE} strategies
for the CSI of a normal, a log-normal, and a Poisson log-normal IFT
signal inference problem and demonstrate their performances via their
\textsc{NIFTy} implementations\textsc{. }
\end{abstract}

\maketitle

\section{Introduction}

\subsection{Correlated signal inference}

Correlated signal fields appear in many contexts as descriptions of
properties of some spatial or temporal domain, like the atmospheric
temperature and velocity, the magnetic field in the Galaxy, a population
density, the evolution of a stock price, etc. The detailed knowledge
of such fields is of high scientific, technological, sociological,
economical, or other interest and rests on data converted into field
estimates. As measurements sample the fields usually in a sparse fashion,
such estimates require interpolations and extrapolations of field
values. Such interpolations are, thanks to the correlation most fields
exhibit, usually justified. However, the optimal reconstruction scheme
requires precise knowledge about the field correlation structure in
order to optimally suppress the measurement noise without over-smoothing
the signal. This correlation structure is usually not known a priori
and has to be inferred along with the signal field. This complicates
the inference substantially, even in case the field statistics is
simply Gaussian. Here we will address this \emph{correlated signal
inference} (CSI) problem within the framework of \textit{information
field theory} (IFT) \citep{2009PhRvD..80j5005E}, the information
theory for fields, which is from a mathematical perspective a statistical
field theory. 

The CSI problem has already been addressed within IFT successfully
by renormalization calculations \citep{2011PhRvD..83j5014E}, which
led to well functioning signal estimators. In order to improve the
accuracy and most importantly the efficiency of such estimators, we
re-investigate their derivation in a slightly different manner than
done before. In particular, the novel approach provides also uncertainty
information on the correlation structure, and takes the impact of
this uncertainty for the field estimate into account.

The problem of estimating the mean of a Gaussian random variable from
independent samples under an unknown covariance was addressed in a
Bayesian setting previously \citep{berger1980,stein1981}. The here
discussed method can be regarded as an extension of those works to
the multidimensional case with independent variables (here field values).
Also this extended problem has been addressed by previous works, most
notably by means of the Gibbs sampling technique \citep{2004ApJS..155..227E,2004PhRvD..70h3511W,2006ApJS..162..401S,2007ApJ...656..653L,2010MNRAS.406...60J}.
This latter approach differs from the here used minimal Gibbs free
energy approach, but shares with it that the field and its spectrum
are inferred simultaneously, and for both uncertainty information
is provided.

\subsection{Structure of this work}

The structure of this work is as follows. Sect.\ \ref{sec:Information-field-theory}
introduces IFT and the Gibbs free energy formalism. Sect.\ \ref{sec:Numerical-information-field}
discusses the challenges of numerical IFT and the \textit{free energy
explorer} (\textsc{FrEE}) strategy to face them. In Sect.\ \ref{sec:Field-knowledge}
all the available information on the fields we investigate here is
mathematically summarized in an information Hamiltonian. In Sect.\ \ref{sec:Renormalization}
the Gibbs free energy is constructed for the normal field measurement
problem. The corresponding log-normal and the Poisson log-normal problems
are treated in Appendices \ref{sec:Log-normal-model} and \ref{sec:Poisson-Log-normal-model},
respectively. The \textsc{FrEE} strategy for the CSI is developed
in Sect.\ \ref{sec:Free-Energy-Exploration} and verified with simulations
in Sect.\ \ref{sec:Examples}. Sect.\ \ref{sec:Conclusion} contains
our conclusions.

\section{Information field theory\label{sec:Information-field-theory}}

\subsection{Basics}

The reconstruction of a physical field from observational data faces
the problem, that the field degrees of freedom outnumber the constraints
given by the data by a huge, if not infinite factor. There are infinitely
many field configurations that are fully consistent with any finite
set of measurements, even in the absence of measurement noise. Most
of those configurations would be discarded as not being plausible.
For example in  many physical fields strong gradients are rare, as
they either require an improbable concentration of energy, or are
rapidly erased by the field evolution. 

To diminish the number of possibilities and to exclude unlikely or
even nonphysical configurations the plausibility of field configurations
should be included in the field reconstruction process. The correct
language to combine measurement information and prior knowledge on
any quantity is probability theory \citep{Cox1963,2003prth.book.....J}. 

IFT is just probability theory for fields. It takes advantage of the
rich pool of methods developed for quantum field theory for the derivation
of optimal field reconstruction methods, which use all available information.
This is achieved by a simple identification of the posterior probability
$\mathcal{P}(\varphi|d,I)$ of a field $\varphi$, constrained by
measurement data $d$ and prior information $I$, with the Gibbs-Boltzmann
distribution of a hypothetical energy function $\mathcal{H}(d,\varphi|I)$
at temperature $\mathcal{T}=1$,

\begin{eqnarray}
\mathcal{P}(\varphi|d,I) & = & \frac{\mathcal{P}(d,\,\varphi|I)}{\mathcal{P}(d|I)}=\frac{e^{-\mathcal{H}(d,\varphi|I)/\mathcal{T}}}{\mathcal{Z}(d|0,I)},\mbox{ with}\nonumber \\
\mathcal{H}(d,\varphi|I) & \equiv & -\ln\mathcal{P}(d,\,\varphi|I)\mbox{ and}\label{eq:Hamiltonian}\\
\mathcal{Z}(d|j,\,I) & \equiv & \mathcal{P}(d|I)=\int\mathcal{D}\varphi\,e^{-\mathcal{H}(d,\varphi|I)/\mathcal{T}+j^{\dagger}\varphi}.\nonumber 
\end{eqnarray}
Here, the field scalar product $j^{\dagger}\varphi=\int dx\,j^{*}(x)\,\varphi(x)$
is used, which we, for notational convenience, assume to be symmetric
by implicitly using $a^{\dagger}b\equiv\frac{1}{2}\left(a^{\dagger}b+b^{\dagger}a\right)$. 

The function $\mathcal{H}(d,\varphi|I)$ therefore summarizes all
information on the field, and is thus called the information Hamiltonian
in IFT. The partition function $\mathcal{Z}(d|j,\,I)$, which is given
by a path integral over all field configurations, is simultaneously
a moment generating function, from which any field moment of interest
can be calculated via derivation with respect to the information source
term $j$. For example the posterior mean is given for $\mathcal{T}=1$
as
\begin{eqnarray}
m & = & \langle\varphi\rangle_{(\varphi|d)}\equiv\int\mathcal{D}\varphi\,\mathcal{\varphi\,P}(\varphi|d)\nonumber \\
 & = & \left.\frac{1}{\mathcal{Z}(d|j)}\,\frac{\delta}{\delta j}\,\int\mathcal{D}\varphi\,e^{-\mathcal{H}(d,\varphi|I)+j^{\dagger}\varphi}\right|_{j=0}\nonumber \\
 & = & \left.\frac{\delta\ln\mathcal{Z}(d|j)}{\delta j}\right|_{j=0}
\end{eqnarray}
and its uncertainty covariance as
\begin{eqnarray}
D & = & \langle(\varphi-m)\,(\varphi-m)^{\dagger}\rangle_{(\varphi|d,\,I)}\nonumber \\
 & = & \left.\frac{\delta^{2}\ln\mathcal{Z}(d|j)}{\delta j\,\delta j^{\dagger}}\right|_{j=0}.
\end{eqnarray}

Unfortunately, the partition function can only be calculated analytically
for a limited class of probability density functionals (PDFs), most
notably the Gaussian distribution 
\begin{equation}
\mathcal{P}(\varphi|\Phi)=\mathcal{G}(\varphi,\Phi)\equiv\frac{1}{\sqrt{|2\pi\Phi|}}\exp\left(-\frac{1}{2}\varphi^{\dagger}\Phi^{-1}\varphi\right),\label{eq:Gaussian}
\end{equation}
where here $\Phi_{xy}=\langle\varphi_{x}\,\varphi_{y}^{\dagger}\rangle_{(\varphi)}$
is the prior field covariance structure. For other PDFs perturbation
theoretic, renormalization, or other approximation techniques have
to be invoked.

The prior Hamiltonian of a Gaussian field with known covariance
\begin{equation}
\mathcal{H}(\varphi|\Phi)=\frac{1}{2}\left(\varphi^{\dagger}\Phi^{-1}\varphi+\ln|2\pi\Phi|\right)\label{eq:free-prior-Hamiltonian}
\end{equation}
is quadratic in $\varphi$, and therefore contributes only to a free
field theory, which can be tackled by linear methods. However, the
CSI problem addressed in this work is how to infer a Gaussian field
$\varphi$ for which the covariance $\Phi$ is not known a priori
and has to be estimated from the same data used to estimate $\varphi$.
This lack of knowledge alone renders an otherwise often easily analytical
solvable problem into a complex, non-Gaussian problem, beyond the
reach of simple perturbation methods.

The available knowledge on the covariance $\Phi$ is summarized by
the hyper-prior $\mathcal{P}(\Phi)$. The prior knowledge on the field
is then
\begin{equation}
\mathcal{P}(\varphi)=\int\mathcal{D}\Phi\,\mathcal{P}(\varphi|\Phi)\,\mathcal{P}(\Phi),
\end{equation}
which turns the corresponding effective Hamiltonian
\begin{eqnarray}
\mathcal{H}(\varphi) & = & -\ln\int\mathcal{D}\Phi\,\mathcal{P}(\varphi|\Phi)\,\mathcal{P}(\Phi)\label{eq:effective_Hamiltonian}\\
 & = & -\ln\int\mathcal{D}\Phi\,e^{-\frac{1}{2}\varphi^{\dagger}\Phi^{-1}\varphi}\,\mathcal{P}(\Phi)+\frac{1}{2}\ln|2\pi\Phi|\nonumber 
\end{eqnarray}
into a non-quadratic function of $\varphi$ unless there is certainty
about the correlation structure, $\mathcal{P}(\Phi)=\delta(\Phi-\Phi_{0})$,
in which case Eq.\ \ref{eq:free-prior-Hamiltonian} is recovered.

All relevant information is stored in the partition function $\mathcal{Z}$.
Any moment of the field can be extracted from $\mathcal{Z}$. However,
the calculation of this can be difficult and therefore approximations
are used to calculate field estimates.

\subsection{Maximum a posteriori}

One popular approximation is the maximum a posteriori (MAP) approximation,
which is equivalent to minimizing the joint Hamiltonian $\mathcal{H}(d,\varphi)$
with respect to the field,
\begin{equation}
\left.\frac{\delta\mathcal{H}(d,\varphi)}{\delta\varphi}\right|_{\varphi=m_{\mathrm{MAP}}}=0.
\end{equation}

Ref.\ \citep{2011PhRvD..83j5014E} showed that this leads to poorly
performing signal reconstruction schemes in case the uncertainty on
$\text{\ensuremath{\Phi}\ is on logarithmic scale. Minimizing }$
$\mathcal{H}(d,\varphi,\Phi)$ simultaneously with respect to $\varphi$
and $\Phi$ gives even worse results. On the MAP level, the only well
working strategy was to estimate $\Phi$ from maximizing $\mathcal{P}(\Phi|d)=\int\mathcal{D\varphi\,}\mathcal{P}(d,\varphi,\Phi)$
and then to use this as given in $\mathcal{P}(d,\varphi|\Phi)$. This
empirical Bayes approach neglects the uncertainties in $\Phi$ and
therefore provides overconfident estimates of $\varphi$.

When using the MAP approach we just need to minimize the Hamiltonian.
This has the advantage that suboptimal steps during the minimization
get corrected by later steps and therefore do not affect the outcome.
A similar guiding function property is desired for a more accurate
estimator of the signal mean. It exists in the form of the Gibbs free
energy.

\subsection{Gibbs free energy}

The Gibbs free energy can be used to approximate a complex posterior
$\mathcal{P}(\varphi|d)$ with a simpler probability function $\widetilde{\mathcal{P}}(\varphi|d)$
\citep{opper2001advanced},\footnote{The minimal Gibbs free energy approach used here is mathematical identical
to the concept of variational inference \citep{2016arXiv160100670B}
based on minimal Kullback-Leibler distance of probability distributions
\citep{Kullback1951}. Given that the oldest of these similar, if
not identical concepts is the Gibbs free energy \citep{Gibbs1873},
we stick to this term in our naming conventions. } for example a Gaussian 
\[
\widetilde{\mathcal{P}}(\varphi|d)=\mathcal{G}(\varphi-m,\,D).
\]
For this, the form
\begin{eqnarray*}
G(m,D|d) & = & U(m,D|d)-\mathcal{T}\,\mathcal{S}(D|d)\\
 & \equiv & \langle\mathcal{H}(d,\varphi)\rangle_{\widetilde{\mathcal{P}}(\varphi|d)}+\mathcal{T}\,\langle\ln\widetilde{\mathcal{P}}(\varphi|d)\rangle_{\widetilde{\mathcal{P}}(\varphi|d)}
\end{eqnarray*}
is most convenient, as its construction requires only Gaussian averaging
to get the internal energy $U$ and the entropy $\mathcal{S}$. The
Gibbs free energy has the desired property of being a guiding objective
function since its minimum is at the posterior mean and in addition
to this its curvature there encodes the uncertainty dispersion:
\begin{eqnarray}
\frac{\delta G}{\delta m}=0\Rightarrow m & = & \langle\varphi\rangle_{(\varphi|d)}\\
\left.\left(\frac{\delta^{2}G}{\delta m\delta m^{\dagger}}\right)^{-1}\right|_{m=\langle\varphi\rangle_{(\varphi|d)}}\!\!\!\!\!\!\!\!\!\!\!\!\!\!\!\!\!\!\!\!\!\!\!\!\mathcal{T}\,\,\, & = & \langle\varphi\,\varphi^{\dagger}\rangle_{(\varphi|d)}.
\end{eqnarray}
The Gibbs free energy was successfully used in Ref.~\citep{2010PhRvE..82e1112E}
to minimize the effective Hamiltonian in Eq.\ \eqref{eq:effective_Hamiltonian}
approximately. To lowest order, the result was identical to that of
the empirical Bayes approach discussed above. In order to obtain a
more accurate field and covariance estimation, including uncertainty
information for both, and hopefully also a numerically more efficient
algorithm, we make a different design choice here. We will not use
the field covariance marginalized effective theory as given by Eq.\ \eqref{eq:effective_Hamiltonian},
but keep the covariance explicit by investigating 
\[
\mathcal{P}(\varphi,\Phi|d,\,I)=\frac{\mathcal{P}(d|\varphi,\,I)\,\mathcal{P}(\varphi|\Phi,\,I)\,\mathcal{P}(\Phi|I)}{\mathcal{P}(d|I)}.
\]
Thus, the unknown ``signal'' to be estimated by the Gibbs formalism
will be the field $\mathcal{\varphi}$ and its prior covariance $\Phi$.
The latter is described for homogeneous statistics by a strictly power
spectrum $P_{\varphi}(k)$. As this function of the Fourier wavevector
$k$ varies often over orders of magnitude, and to enforce its positivity,
we actually use the logarithm of the spectrum in comparison to some
pivot amplitude $r$ as our unknown parameter $\tau_{k}=\ln\left[P_{\varphi}(k)/r\right]$.
Together with the unknown field this forms the combined signal vector
$s=(\varphi^{\dagger},\tau^{\dagger})^{\dagger}$. For the log-power
spectrum a Gaussian approximation of the uncertainty is more appropriate,
as it can be positive or negative equally well.  

\section{Numerical information field theory \label{sec:Numerical-information-field}}

\subsection{Basics}

The equations of IFT have to be solved numerically. These usually
involve high dimensional linear or non-linear operators acting on
fields, which can not be represented explicitly by handling their
matrix elements in computer memory. Instead, these operators have
to be represented implicitly by computer routines that perform the
action of the operator without representing its matrix elements explicitly.
For example the Fourier transformation operator $F$, with $F_{kx}=e^{i\,k\,x}$,
can be represented by the Fast Fourier Transform (FFT) algorithm.
The inverse of such an implicit operators $A$ applied to a vector
$b$ has then to be done by a numerical solver for $A\,x=b$ that
is able to use such implicit operator representations. One such solver
is the famous conjugate gradient (CG) method for solving linear systems
of equations \citep{hestenesstiefel}.

IFT equations usually do not depend explicitly on the size, coordinate
system, number of dimensions, or topology of the space the field of
interest is living on. They look the same for fields over Cartesian
spaces and fields over spheres if the corresponding harmonic spaces
are automatically used to express spatial correlation structures.
In order to benefit from space invariance and in order to facilitate
the implementation of IFT algorithms\textit{ numerical information
field theory} (\textsc{NIFTy}) was developed \citep{2013A&A...554A..26S,2013ascl.soft02013S}.
\textsc{NIFTy} permits the user to specify field inference algorithm
in a coordinate free manner, so that those can easily be applied to
reconstruct fields over varying dimensions and space topologies. It
has successfully been used in a number of imaging and signal inference
problems \citep{2011PhRvE..84d1118O,2012A&A...542A..93O,2013PhRvE..87c2136O,2013PhRvD..88j3516D,2013arXiv1311.5282J,2015A&A...574A..74S,2015PhRvE..91a3311D,2015A&A...581A..59J,2015arXiv151203480G}. 

\textsc{NIFTy} already contains an implementation of the empirical
Bayes CSI, the so called \textit{critical filter} method to retrieve
a field and its unknown spectrum \citep{2011PhRvD..83j5014E,2011PhRvE..84d1118O,2013PhRvE..87c2136O}.
This, however, has three short comings we want to overcome here: 

First, the \textit{critical filter} is relatively slow, as the field
and the spectrum estimates are performed separately by iterations.
These variables have a large inertia since the mutual dependencies
require a combined solution for those. Abstractly speaking, there
exists a diagonal and curved valley in the information Hamiltonian
of combined field and spectrum Hilbert space. This should be followed
downhill, but\textit{ the critical filter} performs only steps in
orthogonal directions of which none is parallel to the valley.

Second, the\textit{ critical filter} has no notion of spectral uncertainty.
Uncertainty of the spectrum should however be manifest in the inference
equations since a field estimator should be more conservative as larger
this uncertainty is.

Third, the\textit{ critical filter} is not adapted to non-linear measurement
situations. It has been used for such, though, but without a strict
IFT derivation.

To address all these issues, but also as a design principle for efficient
and consistent signal inference algorithms with \textsc{NIFTy}, we
introduce here the \textit{free energy exploration} (\textsc{FrEE})
strategy.

\subsection{\textsc{FrEE} strategy}

The \textsc{FrEE} strategy proposes to address signal field inference
problems based on the following design principles:

\textbf{Form follows function:} The optimal numerical scheme should
derive from the properties of a suitable objective function, which
is the Gibbs free energy for a given inference problem. We exploit
the analytic form of this function as much as possible. 

\textbf{Free energy exploration:} The Gibbs energy measures the information
distance of an assumed, approximate probability distribution from
the correct one. Minimizing this distance provides a nearly information
optimal approximate (see \citep{2016arXiv161009018L} for a discussion
of this), from which a good estimate of the posterior mean and uncertainty
can be read off. 

\textbf{Free fields:} A free field follows Gaussian statistics. Information
about field values can then be easily propagated from one location
to another. Consequently, inference of Gaussian fields conditional
to other fields is easier than inference of self-interacting, non-Gaussian
fields. Complex interaction terms in an information Hamiltonian often
result from marginalization of nuisance fields. These are better regarded
as signal fields to be estimated as well than being marginalized.
It can be more efficient numerically to have an explicit representation
of a nuisance field than to recalculate its effective forces from
the signal field in every computational step. If necessary, we can
introduce additional, auxiliary fields to keep track of and/or accumulate
knowledge on field properties that are otherwise expensive to re-calculate.
If our approximate posterior is free, i.e. Gaussian, marginalization
of nuisance parameters can easily be done after the calculation. 

\textbf{Force-free configurations:} The negative gradient of the Gibbs
energy is the force which drives the inference. If, however, a location
for which some of the force components vanish can be reached directly,
we adapt it. 

\textbf{No unnecessary accuracy:} The free energy extremal principle
guides the inference to the optimal solution from any location in
the Hilbert space of the field. Intermediate errors can be corrected
later. If the situation is rapidly changing, there is no need to accurately
plan for the perfect next step or to accumulate statistics for a stochastically
estimated quantity. It is more beneficial to keep moving quickly.

\textbf{Stability:}  The tight coupling between degrees of freedom
of the inference problem, and most notably the concaveness of interesting
problems can easily destabilize a finite step size numerical scheme.
We propose to increase stabilizing ``mass terms'' of the dynamics
for degrees of freedom that exhibit sign changes of their steps. 

In the following, we will construct efficient CSI algorithms based
on the \textsc{FrEE} strategy, while trying to be pragmatic.

\subsection{\textsc{FrEE} recipe}

The recipe to tackle inference problems via the \textsc{FrEE} strategy
is as follows:
\begin{enumerate}
\item Identification of the relevant quantities to be inferred, the signal
$s$ in the following. For the CSI problem this is the combined vector
$s=(\varphi^{\dagger},\tau^{\dagger})^{\dagger}$ describing the field
and its log power spectrum.
\item Construction of $\mathcal{H}(d,s)=-\ln\mathcal{P}(d|s)-\ln\mathcal{P}(s)$,
the joint information Hamiltonian, out of likelihood $\mathcal{P}(d|s)$
and prior $\ln\mathcal{P}(s)$.
\item Choosing an Ansatz for the approximate posterior, here and usually
a Gaussian $\widetilde{\mathcal{P}}(s|d)=\mathcal{G}(s-\overline{s},\,S)$.
The choice for the representation of the posterior uncertainty dispersion
$S$ might require a trade off between accuracy and complexity.
\item Calculation of the internal energy 
\begin{eqnarray*}
U(\overline{s},\,S|d) & = & \langle\mathcal{H}(d,s)\rangle_{\widetilde{\mathcal{P}}(s|d)}\\
 & = & \int\mathcal{D}s\,\mathcal{G}(s-\overline{s},\,S)\,\mathcal{H}(d,s),
\end{eqnarray*}
entropy 
\begin{eqnarray*}
\mathcal{S}(\overline{s},S|d) & = & \int\mathcal{D}s\,\widetilde{\mathcal{P}}(s|d)\,\ln\widetilde{\mathcal{P}}(s|d)\\
 & = & \frac{1}{2}\mathrm{Tr}\left[1+\ln\left(2\pi S\right)\right],
\end{eqnarray*}
and Gibbs free energy 
\[
G(\overline{s},\,S|d)=U(\overline{s},\,S|d)-\mathcal{T}\,\mathcal{S}(\overline{s},S|d),
\]
where $\mathcal{T}=1$.\footnote{For $\mathcal{T}=1$ the Gibbs free energy is -- up to an irrelevant
constant -- identical to the cross information or Kullback-Leibler
distance of the posterior approximation $\widetilde{\mathcal{P}}(s|d)$
to the exact posterior $\mathcal{P}(s|d,\,I)$. For this reason, the
temperature $\mathcal{T}$ will typically be set to the canonical
value of $1$. For $\mathcal{T}=0$ the approximate posterior becomes
a delta function at the maximum of the correct posterior. For problems
with nuisance parameters, as here the field covariance, maximum a
posterior estimates can be very sub-optimal \citep{2011PhRvD..83j5014E}.
We will keep $\mathcal{T}$ in the formula, as it permits the exploration
of inference schemes in between maximum a posteriori ($\mathcal{T}=0$)
and maximal cross information ($\mathcal{T}=1$) or even to increase
the uncertainty estimate in a systematic and self consistent fashion
in case conservative estimates are required ($\mathcal{T}>1$).}
\item Calculation of the information forces 
\[
f_{\overline{s}}=-\frac{\delta G}{\delta\overline{s}}\mbox{ and }f_{S}=-\frac{\delta G}{\delta S}.
\]
 For the latter the force-free solution, $f_{S}=0$, can often be
found analytically and adapted immediately to specify $S$. This might
require the evaluation of trace terms, for which we propose the usage
of stochastic probing (App.\ \eqref{subsec:Auxiliary-fields}).
\item Translation of the forces into an optimal step $\overline{s}\leftarrow\overline{s}+\Delta\overline{s}$
according to a Newton scheme for $\overline{s}$: 
\[
\Delta\overline{s}=-\left(\frac{\delta^{2}G}{\delta\overline{s}\delta\overline{s}^{\dagger}}\right)^{-1}\frac{\delta G}{\delta\overline{s}}=\mathcal{T}^{-1}S\,f_{\overline{s}}.
\]
\item Eventually, modification of $S^{-1}$ in the above step size calculation
to ensure spectral positivity needed for a stable numerical Newton
scheme. This modification should not prevent the scheme to converge
towards the minimum of $G$ as characterized by $f_{\overline{s}}=0$. 
\item Implementation and testing of the resulting \textsc{FrEE} solver.
\end{enumerate}
We will follow this \textsc{FrEE}\textbf{ }recipe in the following
construction of our CSI, and extend it where appropriate. 

\section{Field knowledge\label{sec:Field-knowledge}}

\subsection{Unknown Gaussian field}

A continuous field $\varphi:\Omega\rightarrow\mathbb{K}$ over a metric
space $\Omega$ is assumed to be drawn from a homogeneous Gaussian
random process $\mathcal{P}(\varphi|\Phi)=\mathcal{G}(\varphi,\Phi)$
with harmonic covariance structure $\Phi=\langle\varphi\,\varphi^{\dagger}\rangle_{(\varphi)}$.
The corresponding information Hamiltonian is
\begin{equation}
\mathcal{H}(\varphi|\Phi)\equiv-\ln\mathcal{P}(\varphi|\Phi)=\frac{1}{2}\varphi^{\dagger}\Phi^{-1}\varphi+\frac{1}{2}\ln|2\pi\Phi|.\label{eq:prior_hamilton}
\end{equation}

The signal covariance $\Phi$ is unknown, but we will assume that
it to be harmonic, it commutes with the Beltrami-Laplace operator
$\Delta$. This means for $\Omega=\mathbb{R}^{n}$ that 
\begin{equation}
\Phi_{xy}=\langle\varphi(x)\,\overline{\varphi(y)}\rangle_{(\varphi)}=C_{\varphi}(x-y)
\end{equation}
is translation invariant and only depends on the vector between locations
$x$ and $y$. Furthermore, we will assume statistical isotropy, such
that $\Phi_{xy}=C_{\varphi}(|x-y|)$ only depends on the distance
of $x$ and $y$, but not on their orientation. As a consequence,
the covariance $\Phi$ is characterized by a single 1D function, the
correlation function $C_{\varphi}(r)$, and the covariance is diagonal
in the harmonic or Fourier space, with the power spectrum $P_{\varphi}(k)$
on its diagonal. In the following we parametrize this power spectrum
in a way that is convenient for the \textsc{FrEE} approach.

\subsection{Unknown harmonic correlation structure}

We assume the covariance $\Phi$ to be diagonal in the harmonic basis
of the space $\Omega$ the field $\varphi$ lives over, as this is
the best generalisation of translation invariant statistics. Harmonic
means that the basis functions $a_{k'}:\Omega\mapsto\mathbb{K}$ are
eigenvectors of the Laplace operator $\Delta$. For a Cartesian space
$\Omega=\mathbb{R}^{n}$ this is the Fourier basis, for the sphere
$\Omega=\mathcal{S}^{2}$, this is the basis given by the spherical
harmonic functions. 

Let $\{a_{k'}\}_{k'}$ be a set of orthonormal harmonic eigenvectors
$a_{k'}:\Omega\rightarrow\mathbb{K}$, with corresponding eigenvalues
$\{\lambda_{k'}\}_{k'}$, such that $-\Delta a_{k'}=\lambda_{k'\,}a_{k'}$,
sorted as $\lambda_{k'+1}\ge\lambda_{k'}>\lambda_{0}=0$ for all $k'>0$,
and the $a_{k'}$'s being orthonormal according to $a_{k'}^{\dagger}a_{l'}=\delta_{k'l'}$.
The index $k'$ in principle only labels the different modes, but
for the Fourier space, it can be identified with the Fourier wavevectors
$\vec{k}=\vec{k}(k')$, such that $a_{k'}(\vec{x})=e^{i\vec{k}\cdot\vec{x}}$
and for the spherical harmonic space with the angular momentum quantum
numbers usually (but not here) denoted by $\ell$ and $m$. The reason
for the prime in the notation $k'$ will become clear soon.

Then any harmonic covariance can be expressed as 
\begin{equation}
\Phi=\sum_{k'}p_{k'}a_{k'}a_{k'}^{\dagger}=\sum_{k'}p_{k'}\mathbb{P}_{k'}\equiv p^{\dagger}\mathbb{P},
\end{equation}
where $p=(p_{k'})_{k'}$ is the unknown spectrum of $\Phi$ (the set
of eigenvalues of $\text{\ensuremath{\Phi}}$ ordered by $k'$, the
index of the harmonic operator) and $a_{k'}a_{k'}^{\dagger}\equiv\mathbb{P}_{k'}$
are spectral projectors onto the individual basis vectors. We denote
with $\mathbb{P}=($$\mathbb{P}_{k'})_{k'}$ the vector of all such
spectral projectors. The inverse covariance is then 
\begin{equation}
\Phi^{-1}=\sum_{k'}p_{k'}^{-1}a_{k'}a_{k'}^{\dagger}=\sum_{k'}p_{k'}^{-1}\mathbb{P}_{k'}=p^{-1\dagger}\mathbb{P},
\end{equation}
where we used the notation convention that a function is applied to
a vector component wise in the vector\textquoteright s natural basis.
This also applies to the multiplication or division of vectors, such
as $a\,b=\left(a_{i}b_{i}\right)_{i}$ and $a/b=\left(a_{i}/b_{i}\right)_{i}$.

\subsection{Spectral bands}

It can be convenient to group several harmonic modes together, e.g.
in case their eigenvalues are so similar that a similar spectral amplitude
can be assumed. To this end, we define disjoint and covering sets
of spectral modes $b_{k}=\{k'_{i_{k}},\,k'_{i_{k}+1},\ldots k'_{i_{k}+\varrho_{k}}\}$
with similar harmonic eigenvalues $\{\lambda_{i_{k}},\,\lambda_{i_{k}+1},\ldots\lambda_{i_{k}+\varrho_{k}}\}$
and $\varrho_{k}$ entries. A projector into such spectral bands is
then 
\begin{equation}
\mathbb{P}_{k}=\sum_{k'\in b_{k}}a_{k'}a_{k'}^{\dagger}
\end{equation}
and its trace $\varrho_{k}=\mathrm{Tr}\,\mathbb{P}_{k}$ provides
the number of modes per band $k$. From now on we assume that this
grouping has happened and that every $k$ labels a different $\lambda_{k}$.
Furthermore, if the number $n_{\mathbb{P}}$ of spectral bands is
small the $n_{\mathbb{P}}\times n_{\mathbb{P}}$ matrices can be managed
explicitly in computer memory. We also use the notation $\mathbb{P}=\left(\mathbb{P}_{k}\right)_{k}$
for the vector of spectral projector. This implies, e.g. $\mathbb{P}^{\dagger}1=\sum_{k}\mathbb{P}_{k}=\sum_{k}\sum_{k'\in b_{k}}a_{k'}a_{k'}^{\dagger}=\sum_{k'}a_{k'}a_{k'}^{\dagger}=\mathbb{1}$.
Here and in the following, we use the primed variables $k'$, $l'$,
and $q'$ to denote the original harmonic modes, and the not primed
variables $k$, $l$, and $q$ for harmonic bands.

The spectral vector $p$ parametrizes the covariance $\Phi=\Phi^{(p)}$.
Our a priori knowledge of the covariance can hence be expressed as
$\mathcal{P}(\Phi^{(p)}|I)=\mathcal{P}(p|I)$,\footnote{Here the functional determinant of the transformation $p\rightarrow\Phi^{(p)}$
was taken into account. This is $|\partial\Phi^{(p)}/\partial p|=|\sum_{k}a_{k}a_{k}^{\dagger}|=|\mathbb{1}|=1$
by the orthonormality of the chosen harmonic basis. } where $I$ encodes the set of our a priori assumptions. These should
encode ($i$) the unknown magnitude of the spectral amplitudes and
($ii$) their spatial scale-invariance, such that $i,ii\in I$.

\subsubsection*{$i$. Unknown magnitude of the spectral amplitudes }

Without any prior information on the magnitude of the spectral band
amplitudes $p=(p_{k})_{k}$, these should be modeled as a falling
power law distributions $\mathcal{P}(p_{k}|i)\propto\left(p_{k}/r\right)^{-\alpha_{k}}$
with spectral indexes $\alpha=(\alpha_{k})_{k}$ and reference power
scale $r$. A magnitude-agnostic prior would be given by $\alpha_{k}=1$,
a pessimistic by $\alpha_{k}\ge1$, and an optimistic by $\alpha_{k}\le1$. 

We permit for a lower cut-off of the amplitudes by introducing the
dimensionless parameters $q=(q_{k})_{k}$ and the inverse Gamma-distributions
\begin{equation}
\mathcal{P}(p_{k}|i)\propto\left(\frac{p_{k}}{r}\right)^{-\alpha_{k}}\exp\left[-q_{k}\left(\frac{p_{k}}{r}\right)^{-1}\right].
\end{equation}

Since the spectral magnitudes $\ln\left(p_{k}/r\right)\equiv\tau_{k}$,
play a special role we provide their information Hamiltonian
\begin{eqnarray}
\mathcal{H}(\tau_{k}|i) & = & -\ln\left(\mathcal{P}(p_{k}|i)\,\left\Vert \frac{\partial p_{k}}{\partial\tau_{k}}\right\Vert \right)\nonumber \\
 & \widehat{=} & (\alpha_{k}-1)\,\tau_{k}+q_{k}\,e^{-\tau_{k}}
\end{eqnarray}
 without any irrelevant $\tau_{k}$-independent constants. For later
reference, we state
\begin{equation}
p=r\,e^{\tau}\label{eq:power_convention}
\end{equation}
and note that except for units, $r$ could be absorbed by a redefinition
of $\tau\mapsfrom\tau+\ln r$. This, however, can easily lead to confusion
and therefore we better track $r$ through the formalism.

The Hamiltonian combined for all modes $k$ is
\begin{equation}
\mathcal{H}(\tau|i)\,\widehat{=}\,(\alpha-1)^{\dagger}\tau+q^{\dagger}e^{-\tau},
\end{equation}
 where $\alpha^{\dagger}\tau=\alpha_{k}^{\dagger_{k}}\tau_{k}\equiv\sum_{k}\alpha_{k}\tau_{k}$
denotes the scalar product in the band-harmonic space.

The purely agnostic magnitude prior $i$ with $(\alpha,\tau)=(1,0)$
does not provide any constraints on the magnitudes, since $\mathcal{H}(\tau|i,\mathrm{agnostic})=\mathrm{const}$.

\subsubsection*{$ii$. Spatial scale-invariance of spectral amplitudes}

We further assume that the amplitudes are spatially scale-invariant.
The spatial scale $\mathcal{L}_{k}$ of a harmonic mode $k$ is encoded
in its eigenvalue with respect to $-\Delta$ according to $\lambda_{k}=\mathcal{L}_{k}^{-2}$,
and the magnitude of its inverse as $\mathscr{\kappa}_{k}=\ln\mathcal{L}_{k}^{-1}=\nicefrac{1}{2}\ln\lambda_{k}$.
For Euclidian spaces, $\kappa$ would be the logarithmic length of
the Fourier-vector of a mode.

We postulate the existence of a smooth, but unknown function $\check{\tau}(\kappa)$
such that $\tau_{k}=\check{\tau}(\kappa_{k})$. Smoothness is enforced
by the prior energy
\begin{eqnarray}
\mathcal{H}(\check{\tau}|ii) & \widehat{=} & \int_{-\infty}^{\infty}\frac{d\kappa}{2\sigma^{2}}\,\left(\frac{\partial^{2}\check{\tau}(\kappa)}{\partial\kappa^{2}}\right)^{2}\nonumber \\
 & \equiv & \frac{1}{2}\check{\tau}^{\dagger}\check{T}\,\check{\tau}\approx\frac{1}{2}\tau^{\dagger}T\,\tau\,\widehat{=}\,\mathcal{H}(\tau|ii)\label{eq:Hii}
\end{eqnarray}
where $\sigma$ regulates the strength of the smoothness enforcing.
$\sigma=\infty$ encodes no smoothness, since then $\mathcal{H}(\tau|ii,\,\mathrm{rough})=\mathrm{const}$,
$\sigma=0$ enforces a single power-law spectrum of $p_{k}$, and
$\sigma=1$ seems to be a reasonable compromise between these regimes,
permitting the spectral index of the power spectrum to change typically
by one per $e$-folding in the length scale. 

$T$ represents a discretization of the integro-differential operator
$\check{T}$ onto the spectral grid $\{\kappa_{k}\}_{k}$. A straightforward
discretization $\check{T}\mapsto T$ of the differential operator
$\check{T}=\Delta^{\dagger}\Delta$ in Eq.\ \eqref{eq:Hii} can be
found in Refs.\ \citep{2011PhRvD..83j5014E,2013PhRvE..87c2136O}. 

Note that the zero mode $k=0$ with $a_{0}(x)=\mathrm{const}$, $\lambda_{0}=0$,
and thus $\kappa_{0}=\nicefrac{1}{2}\ln\lambda_{0}=-\infty$ is infinitely
distant to all the other modes, which have $k>0,$ $\lambda_{k}>0$,
and thus $\mathscr{\kappa}_{k}=\nicefrac{1}{2}\ln\lambda_{k}>-\infty$.
Therefore the spectral smoothness prior leaves the zero completely
unconstrained. Its value has to be determined fully from the data
or other prior knowledge. 

\subsubsection*{Combined spectral Hamiltonian}

The combined spectral Hamiltonian is
\begin{align}
\mathcal{H}(\tau|I) & =\mathcal{H}(\tau|i)+\mathcal{H}(\tau|ii)\nonumber \\
 & \widehat{=}\frac{1}{2}\tau^{\dagger}T\,\tau+(\alpha-1)^{\dagger}\tau+q^{\dagger}e^{-\tau},\label{eq:spectral_prior}
\end{align}
where the direct information sum reflects the independence of the
two assumptions $i$ and $ii$.

\subsection{Measurements}

Measurements source our knowledge on the field and its spectrum. Here,
we investigate three measurement situations, the normal field measurement
(N), which is the linear measurement of a Gaussian field as given
by Eq. \eqref{eq:Gaussian}, the log-normal field measurement (LN),
and the Poisson log-normal field measurement (PLN). To focus the discussion,
we investigate the normal case in the main text and treat the LN and
PLN cases in Appendices \ref{sec:Log-normal-model} and \ref{sec:Poisson-Log-normal-model},
respectively.

\subsubsection{Normal field measurement}

A normal field measurement apparatus applies a known linear response
$R$ to the Gaussian field $\varphi$. The response transforms the
continuous field into a discrete data vector to which noise adds,
assumed here to be Gaussian with signal independent and known covariance
$N$. Therefore, we have
\begin{equation}
d\underset{\mathrm{N}}{=}R\,\varphi+n,
\end{equation}
where $\underset{\mathrm{N}}{=}$ is used to indicate that this relation
holds only for the normal (N) field measurement, if this is not noted
directly in the equation. Consequently, the likelihood is (after a
marginalization of the noise)
\begin{equation}
\mathcal{P}(d|\varphi,\,\mathrm{N})=\mathcal{G}(d-R\,\varphi,\,N)
\end{equation}
as well as
\begin{equation}
\mathcal{H}(d|\varphi,\,\mathrm{N})\widehat{=}\frac{1}{2}\left(d-R\,\varphi\right)^{\dagger}N^{-1}\left(d-R\,\varphi\right),\label{eq:L}
\end{equation}
where we dropped as usual unessential constants.

In case the field covariance $\Phi$ is known, the joint Hamiltonian
of data and field is
\begin{eqnarray}
\mathcal{H}(d,\,\varphi|\Phi,\,\mathrm{N}) & = & \mathcal{H}(d|\varphi,\,\mathrm{N})+\mathcal{H}(\varphi|\Phi)\nonumber \\
 & \widehat{=} & \frac{1}{2}\left(d-R\,\varphi\right)^{\dagger}N^{-1}\left(d-R\,\varphi\right)\nonumber \\
 &  & +\frac{1}{2}\varphi^{\dagger}\Phi^{-1}\varphi\nonumber \\
 & \widehat{=} & \frac{1}{2}\left(\varphi-m\right)^{\dagger}D^{-1}\left(\varphi-m\right),
\end{eqnarray}
with 
\begin{equation}
D^{-1}\underset{\mathrm{N}}{=}\Phi^{-1}+R^{\dagger}N^{-1}R\label{eq:D-1}
\end{equation}
the so called posterior precision matrix and 
\begin{equation}
m\underset{\mathrm{N}}{=}D\,R^{\dagger}N^{-1}d\label{eq:WienerMean}
\end{equation}
the posterior mean. $D^{-1}$ is read of from the Hamiltonian and
$m$ is found by quadratic completion. The posterior is then Gaussian
with mean $m$ and uncertainty covariance $D$, 
\begin{equation}
\mathcal{P}(\varphi|d,\,\Phi,\,\mathrm{N})\underset{\mathrm{N}}{=}\mathcal{G}(\varphi-m,\,D).\label{eq:WienerFilter}
\end{equation}

In this case of known covariance $\Phi$, the mean $m$ is a linear
function of the data. The mean $m$ is also called the Wiener filter
solution and $D$ the Wiener variance. In case the field covariance
$\Phi$ is unknown, the linearity between $m$ and $d$ is not the
case any more as we will see later on.

\subsection{Full information Hamiltonian}

The information Hamiltonian of the field and its spectrum
\begin{eqnarray}
\mathcal{H}(d,\,\varphi,\,\tau|I) & = & \mathcal{H}(d|\varphi,\,I)+\mathcal{H}(\varphi|\Phi^{(\tau)},\,I)+\mathcal{H}(\tau|I)\nonumber \\
 & \widehat{\underset{\mathrm{N}}{=}} & \frac{1}{2}\left(d-R\,\varphi\right)^{\dagger}N^{-1}\left(d-R\,\varphi\right)\nonumber \\
 &  & +\frac{1}{2}\varphi^{\dagger}\Phi_{\tau}^{-1}\varphi+\frac{1}{2}\ln|2\pi\Phi_{\tau}|\nonumber \\
 &  & +\frac{1}{2}\tau^{\dagger}T\,\tau+(\alpha-1)^{\dagger}\tau+q^{\dagger}e^{-\tau}\mbox{,}\label{eq:Hamitonian_parts}\\
\text{ with }\Phi_{\tau}^{-1} & = & r^{-1}\,e^{-\tau\dagger}\mathbb{P},\nonumber 
\end{eqnarray}
consists of Eqs. \eqref{eq:Gaussian}, \eqref{eq:spectral_prior},
and \eqref{eq:L}, \eqref{eq:LN}, or \eqref{eq:PLN} for the normal,
log-normal, or Poisson log-normal measurement, respectively. Calculating
the corresponding posterior mean and uncertainty poses a complex problem
that we address approximately by using the \textsc{FrEE} recipe.

\section{Free energy construction \label{sec:Renormalization}}

\subsection{Gaussian Ansatz}

For the construction of the Gibbs free energy we use an approximate
Gaussian Ansatz for the posterior of our signal vector $s^{\dagger}=(\varphi^{\dagger},\tau^{\dagger})$,
consisting of the field an its log-spectrum: 
\begin{eqnarray}
\widetilde{\mathcal{P}}(s|d) & = & \mathcal{G}(s-\overline{s},\,S)\mbox{ with}\label{eq:Gauss_Ansatz}\\
\overline{s} & = & \begin{pmatrix}m\\
t
\end{pmatrix}\mbox{, the posterior mean, and }\nonumber \\
S & = & \begin{pmatrix}D & C\\
C^{\dagger} & \Theta
\end{pmatrix}\mbox{, }\nonumber 
\end{eqnarray}
the posterior uncertainty covariance. 

The posterior mean $\overline{s}$ consists of the mean field $m=(m_{x})_{x}=\langle\varphi\rangle_{\widetilde{\mathcal{P}}}$
and the mean log-spectrum $t=(t_{k})_{k}=\langle\tau\rangle_{\widetilde{\mathcal{P}}}$. 

The signal covariance $S=\langle(s-\overline{s})\,(s-\overline{s})^{\dagger}\rangle_{\widetilde{\mathcal{P}}}$
consists of four blocks: 1) the field uncertainty covariance $D=\langle(\varphi-m)\,(\varphi-m)^{\dagger}\rangle_{\widetilde{\mathcal{P}}}$,
which is a position space operator, $D=(D_{xy})_{xy}$, 2) the log-spectrum
uncertainty covariance $\Theta=\langle(\tau-t)\,(\tau-t)^{\dagger}\rangle_{\widetilde{\mathcal{P}}}$,
which is a harmonic-band space operator, $\Theta=(\Theta_{kl})_{kl}$,
3) the cross correlation $C=\langle(\varphi-m)\,(\tau-t)^{\dagger}\rangle_{\widetilde{\mathcal{P}}}$,
which is an operator that transforms a function over harmonic bands
into one in position space, $C=(C_{xk})_{xk}$, and 4) its adjoined,
$C^{\dagger}=\langle(\tau-t)\,(\varphi-m)^{\dagger}\rangle_{\widetilde{\mathcal{P}}}$,
which transforms from position space into harmonic bands, $C^{\dagger}=(C_{kx}^{\dagger})_{kx}$. 

The unknown parameters $m$, $D$, $t$, $\Theta$, and $C$ depend
in a still to be specified fashion on the data $d$. The Gibbs free
energy is given in terms of these parameters as
\begin{equation}
G(\overline{s},\,S|d)=U(\overline{s},\,S|d)-\mathcal{T}\,\mathcal{S}(\overline{s},\,S|d).\label{eq:Gibbs}
\end{equation}
Here 
\begin{eqnarray}
U(\overline{s},\,S|d) & = & \langle\mathcal{H}(d,\varphi,\tau|I)\rangle_{\widetilde{\mathcal{P}}}\label{eq:U}
\end{eqnarray}
is the internal energy, the full non-Gaussian Hamiltonian of Eq.~\eqref{eq:Hamiltonian}
averaged by the approximate posterior Ansatz Eq.~\eqref{eq:Gauss_Ansatz}.
The entropy of the approximate posterior is 
\begin{eqnarray}
\mathcal{S}(\overline{s},\,S|d) & = & -\!\int\mathcal{\!\!D}s\,\mathcal{P}(s|d,\,I)\,\ln\mathcal{P}(s|d,\,I)\!\!\!\!\nonumber \\
 & = & \langle\mathcal{H}(s|d,\,I)\rangle_{\widetilde{\mathcal{P}}}.\label{eq:S}
\end{eqnarray}

The unknown parameters $m$, $t$, $D$, $\Theta$, and $C$ derive
from a minimization of $G(\overline{s},\,S|d)$ with respect to them.
Ref.~\citep{2010PhRvE..82e1112E} provides implicit formula for this
minimum, under certain approximations which seem to be equivalent
to setting $C=0$ and $\Theta=0$.  These implicit formula have to
be solved numerically. A frequently used strategy in applications
is to iterate the individual implicit formula and to hope for finding
a fix point of the coupled system of equations within an acceptable
computational time. This strategy is used for example in the D$^{3}$PO
code for photon imaging \citep{2015A&A...574A..74S}, the RESOLVE
code for radio interferometry \citep{2013arXiv1311.5282J,2015A&A...581A..59J},
and the tomography code for Galactic reconstruction \citep{2015arXiv151203480G}.
This iterative strategy is computationally expensive as it converges
only slowly.

Here, the minimization of the Gibbs free energy should provide us
directly with an implicit numerical scheme in which the different
equations are solved simultaneously, not iteratively. This is more
efficient and more accurate, if we maintain the spectral uncertainty
information as encoded in $\Theta$ and $C$. 

\subsection{The building blocks }

The Gibbs free energy consists of an entropy term and an internal
energy, which is composed of three terms according to the three Hamiltonian
components in Eq.~\eqref{eq:Hamitonian_parts} resulting from the
likelihood, the field prior, and the spectral hyperprior, respectively. 

\subsubsection{Entropy}

Thanks to the Gaussian Ansatz in Eq. \eqref{eq:Gauss_Ansatz}, the
entropy is independent of $\overline{s}$, 
\begin{align}
\mathcal{S}(\overline{s},\,S|d)= & \frac{1}{2}\mathrm{Tr}\left[1+\ln\left(2\pi S\right)\right]\label{eq:S(S)}
\end{align}
and therefore the gradients of $\mathcal{S}$ with respect to $m$
and $t$ vanish,
\begin{equation}
\frac{\delta\mathcal{S}}{\delta m}=0,\,\frac{\delta\mathcal{S}}{\delta t}=0.
\end{equation}

For later usage, we also provide the gradient of the entropy with
respect to the unknown covariance $S$,
\begin{eqnarray}
\frac{\delta\mathcal{S}}{\delta S} & = & \frac{1}{2}\,S^{-1}\label{eq:S_grad}
\end{eqnarray}
and its sub-covariances $D$, $\Theta$, and $C$. For the latter
we need 
\begin{eqnarray}
S^{-1} & = & \begin{pmatrix}E^{-1} & -D^{-1}C\,F^{-1}\\
-\Theta^{-1}C^{\dagger}E^{-1}\, & F^{-1}
\end{pmatrix}\mbox{, with}\nonumber \\
E & \equiv & D+C\,\Theta^{-1}C^{\dagger}\mbox{ and}\label{eq:S^-1}\\
F & \equiv & \Theta+C^{\dagger}D^{-1}C.\nonumber 
\end{eqnarray}
In case $C=0$, we would have $E=D$ and $F=\Theta$, but otherwise,
the matrix inversion mixes between $D$ and $\Theta$. 

The entropy-gradients with respect to the unknown sub-covariances
are 

\begin{eqnarray}
\frac{\delta\mathcal{S}}{\delta D} & = & \frac{\delta\mathcal{S}}{\delta S}\cdot\frac{\delta S}{\delta D}=E^{-1}=\left(D+C\,\Theta^{-1}C^{\dagger}\right)^{-1},\nonumber \\
\frac{\delta\mathcal{S}}{\delta\Theta} & = & \frac{\delta\mathcal{S}}{\delta S}\cdot\frac{\delta S}{\delta\Theta}=F^{-1}=\left(\Theta+C^{\dagger}D^{-1}C\right)^{-1},\label{eq:Entropic forces}\\
\frac{\delta\mathcal{S}}{\delta C} & = & \frac{\delta\mathcal{S}}{\delta S}\cdot\frac{\delta S}{\delta C}=-E^{-1}C\,\Theta^{-1},\mbox{ and}\nonumber \\
\frac{\delta\mathcal{S}}{\delta C^{\dagger}} & = & \frac{\delta\mathcal{S}}{\delta S}\cdot\frac{\delta S}{\delta C^{\dagger}}=-F^{-1}C^{\dagger}\,D^{-1}.\nonumber 
\end{eqnarray}
Here, $A\cdot B=\mathrm{Tr}\left(A^{\dagger}\,B\right)$ is the matrix
scalar product. We expect 
\begin{equation}
\left(\frac{\delta\mathcal{S}}{\delta C}\right)^{\dagger}=\frac{\delta\mathcal{S}}{\delta C^{\dagger}}
\end{equation}
and therefore 
\begin{eqnarray}
E^{-1}C\,\Theta^{-1} & = & D^{-1}C\,F^{-1},
\end{eqnarray}
which is easily verified by inserting the definitions of $E$ and
$F$ in Eq.\ \eqref{eq:S^-1}. 

The entropic forces $f_{D}^{\mathcal{S}}=-\delta\mathcal{S}/\delta D$
and $f_{\Theta}^{\mathcal{S}}=-\delta\mathcal{S}/\delta\Theta$ ensure
for $\mathcal{T}>0$ non-vanishing uncertainty covariances, as they
push for larger covariances as one can read off from 
\begin{eqnarray}
f_{D}^{\mathcal{S}}\cdot D & = & \mathcal{T\,}\mathrm{Tr}\left[\left(\mathbb{1}+D\,C\,\Theta^{-1}C^{\dagger}\right)^{-1}\right]>0\mbox{ and}\nonumber \\
f_{\Theta}^{\mathcal{S}}\cdot\Theta & = & \mathrm{\mathcal{T\,}Tr}\left[\left(\mathbb{1}+\Theta\,C^{\dagger}D^{-1}C\right)^{-1}\right]>0.
\end{eqnarray}
In contrast, the entropic force $f_{C}^{\mathcal{S}}=-\delta\mathcal{S}/\delta C^{\dagger}$
reduces the magnitude of $C$ as 
\begin{equation}
f_{C}^{\mathcal{S}}\cdot C=-\mathcal{T\,}\mathrm{Tr}\left[\Theta^{-1}C^{\dagger}E^{-1}C\right]\le0.
\end{equation}
All these forces try to increase the entropy. If not counterbalanced
by other forces, they would lead to a state of complete lack of certainty,
$D=\infty$, $\Theta=\infty$, and $C=0$.

\subsubsection{Hyperprior}

The spectral prior $\mathcal{P}(\tau)$ is the hyperprior of our problem.
Its internal energy is
\begin{eqnarray}
U_{\mathrm{\tau}}(\overline{s},\,S|d) & = & \langle\mathcal{H}(\tau)\rangle_{\widetilde{P}}\nonumber \\
 & \widehat{=} & \frac{1}{2}t^{\dagger}T\,t+\frac{1}{2}\mathrm{Tr}\left(T\,\Theta\right)\label{eq:Uh}\\
 &  & +(\alpha-1)^{\dagger}t+q^{\dagger}e^{-t+\widetilde{\Theta}/2}.\nonumber 
\end{eqnarray}
Here and elsewhere, a dropped $\dagger$ between two vectors indicates
component wise multiplication, $\left(q\,t\right)_{k}=q_{k}t_{k}$,
a tilde on a tensor means its diagonal vector in band harmonic basis,
$\widetilde{\Theta}_{k}=\Theta_{kk}$, and a tilde on a vector denotes
a tensor with the vector on the diagonal in band harmonic basis, $\widetilde{t}_{kl}=\delta_{kl}t_{k}$.
As a consequence of this notation we find $\widetilde{\widetilde{t}}=t$
always, but $\widetilde{\widetilde{\Theta}}=\Theta$ only for $\Theta$
being diagonal in the harmonic basis. Similarly we define $\widehat{m}_{xy}=\delta_{xy}\,m_{x}$
as an operator which is diagonal in position basis and $\widehat{D}_{x}=D_{xx}$
the diagonal of $D$ in position basis. 

The corresponding non-vanishing gradients of the hyperprior internal
energy are 
\begin{eqnarray}
\frac{\delta U_{\mathrm{\tau}}}{\delta t^{\dagger}} & = & \frac{1}{2}\,\left[T\,t+(\alpha-1)-q\,e^{-t'}\right],\nonumber \\
\frac{\delta U_{\mathrm{\tau}}}{\delta\Theta} & = & \frac{1}{2}\,\left[T+\widetilde{q\,e^{-t'}}\right],\mbox{ with}\label{eq:Uh_grad}\\
t' & \equiv & t-\widetilde{\Theta}/2.\nonumber 
\end{eqnarray}

\subsubsection{Prior\label{subsec:Prior_inner_energy}}

The field prior $\mathcal{P}(\varphi|\Phi_{\tau})$ provides the internal
energy
\begin{eqnarray}
U_{\mathrm{\varphi}}(\overline{s},\,S|d) & = & \langle\mathcal{H}(\varphi|\Phi_{\tau},\,I)\rangle_{\widetilde{P}}\nonumber \\
 & \widehat{=} & \frac{1}{2}\,\varrho^{\dagger}\mathrm{t}+\frac{1}{2r}\sum_{k}e^{-t'_{k}}\,w_{k},\mbox{with}\label{eq:Uphi}\\
w_{k} & \equiv & \mathrm{Tr}\left\{ \mathbb{P}_{k}\left[\left(m-C_{k}\right)\,\left(m-C_{k}\right)^{\dagger}+D\right]\right\} ,\nonumber 
\end{eqnarray}
and $C_{k}=(C_{kx})_{x}$.\footnote{This result required the calculation of 
\begin{eqnarray*}
\langle e^{-\tau}\varphi\varphi^{\dagger}\rangle_{\widetilde{\mathcal{P}}} & = & \langle e^{-t-\delta\tau}(m+\delta\varphi)\,(m+\delta\varphi)^{\dagger}\rangle_{\widetilde{\mathcal{P}}}\\
 & = & e^{-t}\langle e^{-\delta\tau}\rangle_{\widetilde{\mathcal{P}}}m\,m^{\dagger}\\
 &  & +e^{-t}m\langle e^{-\delta\tau}\,\delta\varphi^{\dagger}\rangle_{\widetilde{\mathcal{P}}}\\
 &  & +e^{-t}\langle e^{-\delta\tau}\,\delta\varphi\rangle_{\widetilde{\mathcal{P}}}m^{\dagger}\\
 &  & +\langle e^{-\delta\tau}\delta\varphi\,\delta\varphi^{\dagger}\rangle_{\widetilde{\mathcal{P}}}\\
 & = & e^{-t'}\left(m\,m^{\dagger}-m\,C^{\dagger}-C\,m^{\dagger}+D+C\,C^{\dagger}\right)\\
\mbox{ with }t' & \equiv & t-\frac{1}{2}\widetilde{\Theta}
\end{eqnarray*}
by usage of a Taylor expansion of $e^{-\delta\tau}$, of the Wick
theorem, and by a re-summation of the resulting series providing the
$e^{\frac{1}{2}\widetilde{\Theta}}$ term. Alternatively, the operator
formalism of Ref.\ \textbf{\citep{2016arXiv160500660L}} can be used
to obtain this result. } The term $\frac{1}{2}\,\varrho^{\dagger}\mathrm{t}$ results from
the log-determinant $\ln|2\pi\Phi|$ in $\mathcal{H}(\varphi|\Phi,I)$.
The multiplicity $\varrho_{k}=\mathrm{Tr}\,\mathbb{P}_{k}$ of a spectral
band $k$ sharing the same harmonic eigenvalue $\lambda_{k}$ enters
since the different harmonic modes count separately in the determinant
in Eq. \eqref{eq:prior_hamilton}. This internal energy would simplify
in case $C=0$ to

\begin{eqnarray}
U_{\mathrm{\varphi}}(\overline{s},\,S|d) & \widehat{=} & \frac{1}{2}\,\left[\varrho^{\dagger}\mathrm{t}+m^{\dagger}\Phi_{\tau}^{-1}m+\mathrm{Tr}\left(\Phi_{\tau}^{-1}D\right)\right]\nonumber \\
\mbox{with }\Phi_{t'}^{-1} & \equiv & r^{-1}\sum_{k}\mathbb{P}_{k}e^{-t'_{k}}\equiv r^{-1}\mathbb{P}^{\dagger}e^{-t'}.\label{eq:Phi}
\end{eqnarray}
In the general case of $C\neq0,$ the associated gradients are
\begin{eqnarray}
\frac{\delta U_{\mathrm{\varphi}}}{\delta m^{\dagger}} & = & \frac{1}{2}\,\Phi_{t'}^{-1}m-\frac{1}{2r}\sum_{k}e^{-t'_{k}}\,\mathbb{P}_{k}\,C_{k},\label{eq:Up_grad}\\
\frac{\delta U_{\mathrm{\varphi}}}{\delta t} & = & \frac{1}{2}\,\left(\varrho-r^{-1}w\,e^{-t'}\right),\nonumber \\
\frac{\delta U_{\mathrm{\varphi}}}{\delta D} & = & \frac{1}{2}\,\Phi_{t'}^{-1},\nonumber \\
\frac{\delta U_{\mathrm{\varphi}}}{\delta\Theta_{kl}} & = & \frac{\delta_{kl}}{4}\,r^{-1}e^{-t'_{k}}w_{k},\mbox{ and}\nonumber \\
\frac{\delta U_{\mathrm{\varphi}}}{\delta C_{k}^{\dagger}} & = & \frac{1}{2r}\,e^{-t'_{k}}\mathbb{P}_{k}\left(C_{k}-m\right).\nonumber 
\end{eqnarray}
It is interesting that $C$ can diminish the $m$-gradient generated
by the prior precision matrix $\Phi_{t'}^{-1}$. Knowing that a more
extreme $m$ will imply a higher inferred spectrum and therefore a
smaller prior precision matrix, the system reduces the restoring force
of the currently assumed spectrum. The last equation states, that
in the absence of any other information forces, $C$ would adapt such
that $\mathbb{P}_{k}C_{k}=\mathbb{P}_{k}m$, which is fulfilled by
$C_{k}^{\text{\textdegree}}=\mathbb{P}_{k}m$ as $\mathbb{P}_{k}\mathbb{P}_{k}=\mathbb{P}_{k}$.
$C=C^{\text{\textdegree}}$ would remove any restoring forces on $m$
from the precision matrix $\Phi_{t'}^{-1}$. There is, however, an
entropic force on $C$ given by Eq.\ \eqref{eq:Entropic forces},
which counteracts the pull towards $C^{\text{\textdegree}}$ and ensures
that the prior precision matrix influences the field inference to
an appropriate degree. 

\subsubsection{Likelihood }

The internal energy of the likelihood term should be denoted by $U_{\mathrm{L}}$
with $\mathrm{L}\in\{\mbox{N, LN, PLN}\}$ for the three likelihoods
under investigation here. For the linear measurement of a normal field
we have 
\begin{eqnarray}
U_{\mathrm{N}}(\overline{s},\,S|d) & = & \langle\mathcal{H}(d|\varphi,\,\mathrm{N})\rangle_{(s|d,\,I)}\nonumber \\
 & \widehat{=} & \mathrm{\frac{1}{2}\,Tr}\left[\left(m\,m^{\dagger}+D\right)\,R^{\dagger}N^{-1}R\right]\nonumber \\
 &  & -m^{\dagger}R^{\dagger}N^{-1}d,\nonumber \\
\frac{\delta U_{\mathrm{N}}}{\delta m} & = & R^{\dagger}N^{-1}\left(R\,m-d\right),\label{eq:UN_grad}\\
\frac{\delta U_{\mathrm{N}}}{\delta D} & = & \frac{1}{2}\,R^{\dagger}N^{-1}R,\nonumber 
\end{eqnarray}
and all other gradients vanish. The corresponding results for $U_{\mathrm{LN}}$
and $U_{\mathrm{PLN}}$ can be found in Appendices \ref{sec:Log-normal-model}
and \ref{sec:Poisson-Log-normal-model}, respectively. The likelihood
provides a pull on the mean field towards $m^{\text{\textdegree}}=(R^{\dagger}N^{-1}R)^{-1}R^{\dagger}N^{-1}d$,
where $(R^{\dagger}N^{-1}R)^{-1}$denotes the pseudo inverse of $R^{\dagger}N^{-1}R$. 

\subsection{Gibbs free energy}

The Gibbs free energy $G=G(\overline{s},\,S|d)$ is then
\begin{eqnarray}
G & \,\widehat{=}\, & U_{\mathrm{L}}+\frac{1}{2}\,\varrho^{\dagger}\mathrm{t}+\frac{1}{2r}\sum_{k}e^{-t'_{k}}\,w_{k}\nonumber \\
 &  & +\frac{1}{2}t^{\dagger}T\,t+\frac{1}{2}\mathrm{Tr}\left(T\,\Theta\right)+(\alpha-1)^{\dagger}t+q\,e^{-t'}\nonumber \\
 &  & -\frac{\mathcal{T}}{2}\,\mathrm{Tr}\left[\ln\left(S\right)\right],
\end{eqnarray}
with $\mathrm{L}\in\{\mbox{N, LN, PLN}\}$ and in particular

\begin{eqnarray}
U_{\mathrm{L}} & \,\widehat{\underset{\mathrm{N}}{=}}\, & \mathrm{\frac{1}{2}\,Tr}\left[\left(m\,m^{\dagger}+D\right)\,R^{\dagger}N^{-1}R\right]\nonumber \\
 &  & -m^{\dagger}R^{\dagger}N^{-1}d.
\end{eqnarray}

\section{\label{sec:Free-Energy-Exploration}Free Energy Exploration}

\subsection{FrEE dynamics}

The solution of the inference problem should be driven by the Gibbs
free force. If $G(\overline{s},\,S|d)$ is the Gibbs free energy of
an inference problem, its negative gradient 
\begin{equation}
f=\begin{pmatrix}f_{\overline{s}}\\
f_{S}
\end{pmatrix}=-\begin{pmatrix}\frac{\delta G}{\delta\overline{s}}\\
\frac{\delta G}{\delta S}
\end{pmatrix}
\end{equation}
in terms of the signal mean $\overline{s}$ and its uncertainty dispersion
$S$ should be the force which drives the dynamics. These forces will
now be calculated for the CSI problem of the normal field measurement
situation. The corresponding forces are composed according to Eq.\ \eqref{eq:Gibbs}
from Eqs.\ \eqref{eq:S_grad}, \eqref{eq:Uh_grad}, \eqref{eq:Up_grad},
and \eqref{eq:UN_grad}.

\subsection{Signal mean}

The mean field evolution is driven by the force
\begin{equation}
f_{m}\underset{\mathrm{N}}{=}R^{\dagger}N^{-1}\left(d-R\,m\right)-r^{-1}\sum_{k}\mathbb{P}_{k}e^{-t'_{k}}(m-C_{k})
\end{equation}
and stops evolving for a static $\Phi$ as soon as
\begin{eqnarray}
m & \underset{\mathrm{N}}{=} & (\Phi_{t'}^{-1}+R^{\dagger}N^{-1}R)^{-1}R^{\dagger}N^{-1}\times\nonumber \\
 &  & \left(d-r^{-1}\sum_{k}\mathbb{P}_{k}e^{-t'_{k}}C_{k}\right)
\end{eqnarray}
 is reached. This is the Wiener filter solution in case $C=0$. However,
$\Phi_{t'}^{-1}=r^{-1}\sum_{k}\mathbb{P}_{k}e^{-t'_{k}}$ evolves
as well due to the force 
\begin{eqnarray}
f_{t} & = & \theta-\left(\alpha-1+\frac{\varrho}{2}+T\,t\right),\mbox{ with}\label{eq:t_dot}\\
\theta & \equiv & \left\{ q+\frac{w}{2\,r}\right\} \,e^{-t'}.\nonumber 
\end{eqnarray}
This stops evolving once 
\begin{equation}
e^{t'_{k}}=\frac{q_{k}+\frac{1}{2r}\,\mathrm{Tr}\left[\left(\left(m-C\right)\left(m-C\right)^{\dagger}+D\right)\,\mathbb{P}_{k}\right]}{\left(\alpha-1+\frac{\varrho}{2}+T\,t\right)_{k}}.\label{eq:t_fix_point}
\end{equation}
This formula reduces to the critical filter formula developed in Refs.
\citep{2011PhRvD..83j5014E,2010PhRvE..82e1112E,2011PhRvE..84d1118O,2013PhRvE..87c2136O}
in case $C=0$ and $\widetilde{\Theta}=0$, the latter implying $t=t'$.
For a fixed $m$, $D$, $\widetilde{\Theta}$, and $C$ this implicit
formula can be solved e.g. by iteration. However, these quantities
evolve themselves and therefore it is advisable to solve the dynamics
of all these quantities simultaneously.

Note, that while $t'=t-\frac{1}{2}\widetilde{\Theta}$ appears as
the effective log spectrum in $\Phi_{t'}^{-1}$ the dynamics is for
$t$ and also the smoothness enforcing term $T\,t$ does not involve
$\widetilde{\Theta}$. The required quantity $w$ needs that $\widetilde{D}=\mathrm{Tr}\left(\mathbb{P}\,D\right)$
as well as $C$ are available according to Eq.\ \eqref{eq:Uphi}.
Also this requests that we investigate the uncertainty covariances
$D$, $C$, and $\Theta$ next. 

\subsection{Uncertainty covariances}

\subsubsection{Field uncertainty covariance}

The field uncertainty covariance force

\begin{equation}
f_{D}\underset{\mathrm{N}}{=}\frac{1}{2}\,\left[\mathcal{T}\,\left(D+C\,\Theta^{-1}C^{\dagger}\right)^{-1}-\left(\Phi_{t'}^{-1}+R^{\dagger}N^{-1}R\right)\right]\label{eq:D_dot}
\end{equation}
vanishes -- for fixed $\Phi_{t'}^{-1}$, $C$, and $\Theta$ --
for the stationary force-free solution

\begin{equation}
D=D^{\text{\textdegree}}\underset{\mathrm{N}}{=}\underbrace{\mathcal{T}\,\left(\Phi_{t'}^{-1}+R^{\dagger}N^{-1}R\right)^{-1}}_{=E}-\underbrace{C\,\Theta^{-1}C^{\dagger}}_{\equiv A_{D}}.\label{eq:D_fix_point}
\end{equation}
It is reasonable to adapt this solution instantaneously at every time
step. First, in most applications, there is no hope to store $D$
in a computer memory. Second, adapting the force-free solution for
$D$ will immediately put us closer to the minimum of the Gibbs free
energy $G$ and therefore speed up the inference.

The term $E=\mathcal{T}\,\left(\Phi_{t'}^{-1}+R^{\dagger}N^{-1}R\right)^{-1}$
to which $D$ reduces for $C=0$ was encountered already in Eq.\ \eqref{eq:D-1}
for the natural choice $\mathcal{T}=1$.\footnote{The extra term $A_{D}=C\,\Theta^{-1}C^{\dagger}$ seems to reduce
the field uncertainty covariance $D$ with respect to $E$ and the
value $D$ has in case of known spectrum (Eq.\ \eqref{eq:D-1}).
However, this conclusion is not correct, as the effective spectrum
$r\,e^{t'}$ appearing in $D$ and $E$ is also decreased in the presence
of $C\sim\mathbb{P}m$, as an inspection of $w$ as given by Eq.\ \eqref{eq:Uphi}
reveals. Thus, this extra term just compensates for some otherwise
too strong enlargement of the field uncertainty covariance caused
by the reduced inferred effective spectrum.} For this term, an implicit representation is possible, as all operator
terms in $E^{-1}$ can be represented as computer routines, and the
application of $E$ to a vector $j$, $m=E\,j$, can therefore be
obtained from a CG based solving of $E^{-1}m=j$. If $C$ and $\Theta$
are available, $D$ can therefore be applied to any vector. For later
reference, we note, that 
\begin{eqnarray}
D^{-1} & = & \left(E-A_{D}\right)^{-1}\nonumber \\
 & = & \left(\mathbb{1}-E^{-1}A_{D}\right)^{-1}E^{-1}
\end{eqnarray}
can as well be represented by an implicit routine, based on a CG inversion
of $\mathbb{1}-E^{-1}A_{D}$. A numerical necessity is that $E^{-1}C\,\Theta^{-1}C^{\dagger}<\mathbb{1}$
all the time and therefore that $\Theta$ stays sufficiently large.

The impact of implicitly solving $\frac{\delta G}{\delta D}=0$ for
$D$ has to be taken into account by adding the effect any changing
quantity affecting $D$ has on $G$. Relevant quantities are $t$,
$C$, and $\Theta$, since those affect $D$ according to Eqs.\ \eqref{eq:D_fix_point},
\eqref{eq:Phi}, and \eqref{eq:Uh_grad}. However, we find the relevant
gradients are not changed, as the implicit solution for $D$ is at
a (constrained) minimum of $G$:
\begin{eqnarray}
\left.\frac{\delta G}{\delta t_{k}}\right|_{D=D^{\text{\textdegree}}} & \equiv & \underbrace{\frac{\delta G}{\delta D}}_{=0}\cdot\frac{\delta D}{\delta t_{k}}=0\mbox{ as well as}\nonumber \\
\left.\frac{\delta G}{\delta\Theta}\right|_{D=D^{\text{\textdegree}}} & = & 0\mbox{ for the same reason,}
\end{eqnarray}
and so forth.

\subsubsection{Spectral uncertainty covariance}

The uncertainty covariance of the spectral field $\tau$has the force
\begin{eqnarray}
f_{\Theta} & = & \frac{1}{2}\,\left[\mathcal{T}\,\left(\Theta+C^{\dagger}D^{-1}C\right)^{-1}-\left(T+\widetilde{\theta}\right)\right],
\end{eqnarray}
acting on it, which drives the $\Theta$-evolution until it reaches
a stable configuration at
\begin{equation}
\Theta^{\text{\textdegree}}=\underbrace{\mathcal{T}\,\left(T+\widetilde{\theta}\right)^{-1}}_{=F}-\underbrace{C^{\dagger}D^{-1}C}_{\equiv A_{\Theta}}.
\end{equation}
We should again adapt the fix point solution $\Theta=\Theta^{\text{\textdegree}}$
immediately at every step of the dynamics. No other dynamical equation
has to be changed in this case for exactly the same reason as the
one we encountered when adapting the temporary fix points for $D$.

For most application, it should be possible to invert $T+\widetilde{\theta}$
numerically at every time step, so that an explicit representation
of $F=\mathcal{T}\,\left(T+\widetilde{\theta}\right)^{-1}$ is available.
Its contribution to $\widetilde{\Theta}$ can be read off, however,
the contribution from the term $A_{\mathrm{\Theta}}=C^{\dagger}D^{-1}C$
is more tricky, as it contains $D^{-1}=\left(E-C\,\Theta^{-1}C^{\dagger}\right)^{-1}$
that again depends on $\Theta$. We propose to shortcut this infinite
recursion by using the following approximations:
\begin{eqnarray}
A_{D} & = & C\,\Theta^{-1}C^{\dagger}\approx C\,F^{-1}C^{\dagger}\nonumber \\
A_{\Theta} & = & C^{\dagger}D^{-1}C\approx C^{\dagger}E^{-1}C\label{eq:A_D_approx}
\end{eqnarray}

\subsubsection{Uncertainty covariance of field and spectrum}

The $C$-force is 
\begin{equation}
f_{C}=\frac{1}{2r}\,e^{-t'}\left[\mathbb{P}\left(m-C\right)\right]-\frac{\mathcal{T}}{2}\,E^{-1}C\,\Theta^{-1},\label{eq:fC}
\end{equation}

Setting the force $f_{C}$ to zero, we can solve for the operator
$C$. To do this, we change into the orthonormal eigenbasis of the
$\Theta$ operator, with eigenvalues $\lambda_{i}$ and corresponding
eigenvectors $a_{i}$, where we can write:

\begin{equation}
\Theta^{-1}=\sum_{i}\lambda_{i}^{-1\,}a_{i}\,a_{i}^{\dagger}
\end{equation}
The operator $C$ can be expressed as:

\begin{equation}
C=\sum_{i}c_{i}\,a_{i}^{\dagger}\label{eq:C_representation}
\end{equation}
Multiplying the force $f_{C}$ by the eigenvector $a_{i}$ gives us
the force on the map $c_{i}$:

\begin{eqnarray}
f_{C}\,a_{i} & = & \dfrac{1}{2r}\,e^{-t'}\,\big[\mathbb{P}(m\,a_{i}-\sum_{j}c_{j}\,a_{j}^{\dagger}\,a_{i})\big]\nonumber \\
 &  & -\frac{1}{2}\mathcal{T}\,E^{-1}\sum_{j}c_{j}\,a_{j}^{\dagger}\,\sum_{l}\lambda_{l}^{-1}\,a_{l}\,a_{l}^{\dagger}\,a_{i}
\end{eqnarray}
Setting this equation to zero and using orthonormality of the eigenvectors,
we get:

\begin{equation}
0\stackrel{!}{=}\dfrac{1}{2r}\,e^{-t'}\,\big[\mathbb{P}(m\,a_{i}-c_{i})\big]-\frac{1}{2\lambda_{i}}\mathcal{T}\,E^{-1}\,c_{i}
\end{equation}

\begin{equation}
\dfrac{1}{2r}\,e^{-t'}\,\mathbb{P}\,m\,a_{i}=\big[\dfrac{1}{2r}\,e^{-t'}\,\mathbb{P}+\frac{1}{2\lambda_{i}}\,\mathcal{T}\,E^{-1}\big]\,c_{i}
\end{equation}
This can be solved and rewritten as
\begin{equation}
c_{i}=\frac{\lambda_{i}}{r}\big[(1+\frac{\lambda_{i}}{r})\Phi^{-1}+R^{\dagger}N^{-1}R\,\big]^{-1}\Phi^{-1}\,a_{i}^{\dagger}\mathbb{P}\,m\label{eq:c_i}
\end{equation}
the force free operator $C$ can now be constructed out of all maps
$c_{i}$ with corresponding eigenvectors $a_{i}$. Calculating the
whole off-diagonal operator involves the numerical inversion of an
operator for each band $i$. For large problems, this computationally
expensive task is impracticable and slows down the inference. Therefore
we propose using only the $c_{i}$ corresponding to the largest eigenvalues
$\lambda_{i}$ of $\Theta$, or to drop $C$ completely. 

\subsubsection{Precision matrix}

The global signal posterior precision matrix can now be written as

\begin{eqnarray}
S^{-1} & = & \begin{pmatrix}E^{-1} & -E^{-1}C\,\Theta^{-1}\\
-\Theta^{-1}C^{\dagger}E^{-1}\, & F^{-1}
\end{pmatrix}\mbox{, with}\nonumber \\
E^{-1} & = & \mathcal{T}^{-1}\left(\Phi_{t'}^{-1}+R^{\dagger}N^{-1}R\right)\mbox{ and}\label{eq:S^-1-1}\\
F^{-1} & = & \mathcal{T}^{-1}\left(T+\widetilde{\theta}\right)\nonumber 
\end{eqnarray}
being well applicable and therefore also numerically invertible operators.
The non-diagonal blocks were written in a form that avoids the operator
$D^{-1}$ and particularly ensures that $S^{-1}$ is always self-adjoined. 

\subsection{Fixing the fix point\label{subsec:Fixing-the-fix}}

\begin{figure*}[t]
\includegraphics[width=0.5\textwidth]{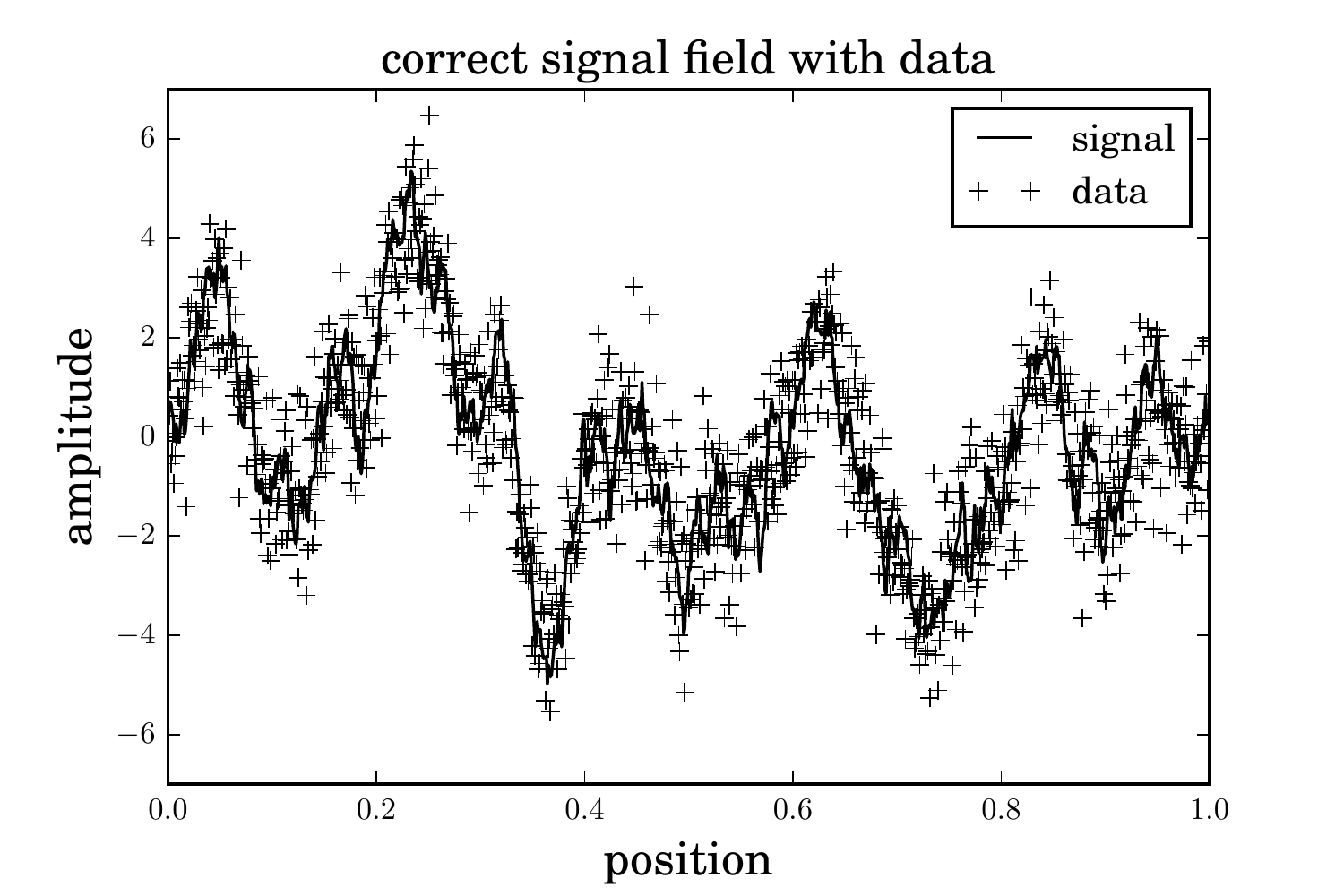}\includegraphics[width=0.5\textwidth]{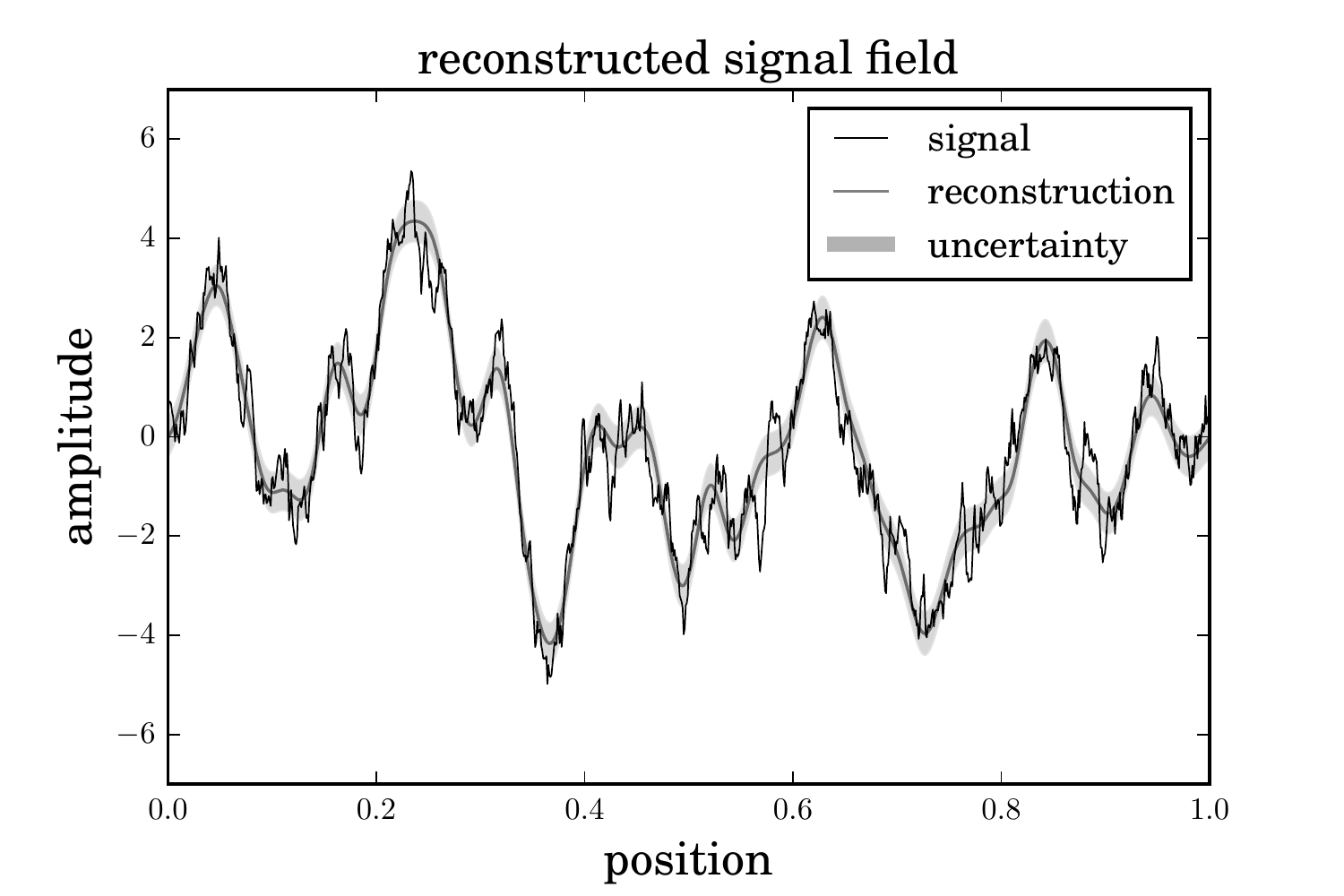}

\includegraphics[width=0.5\textwidth]{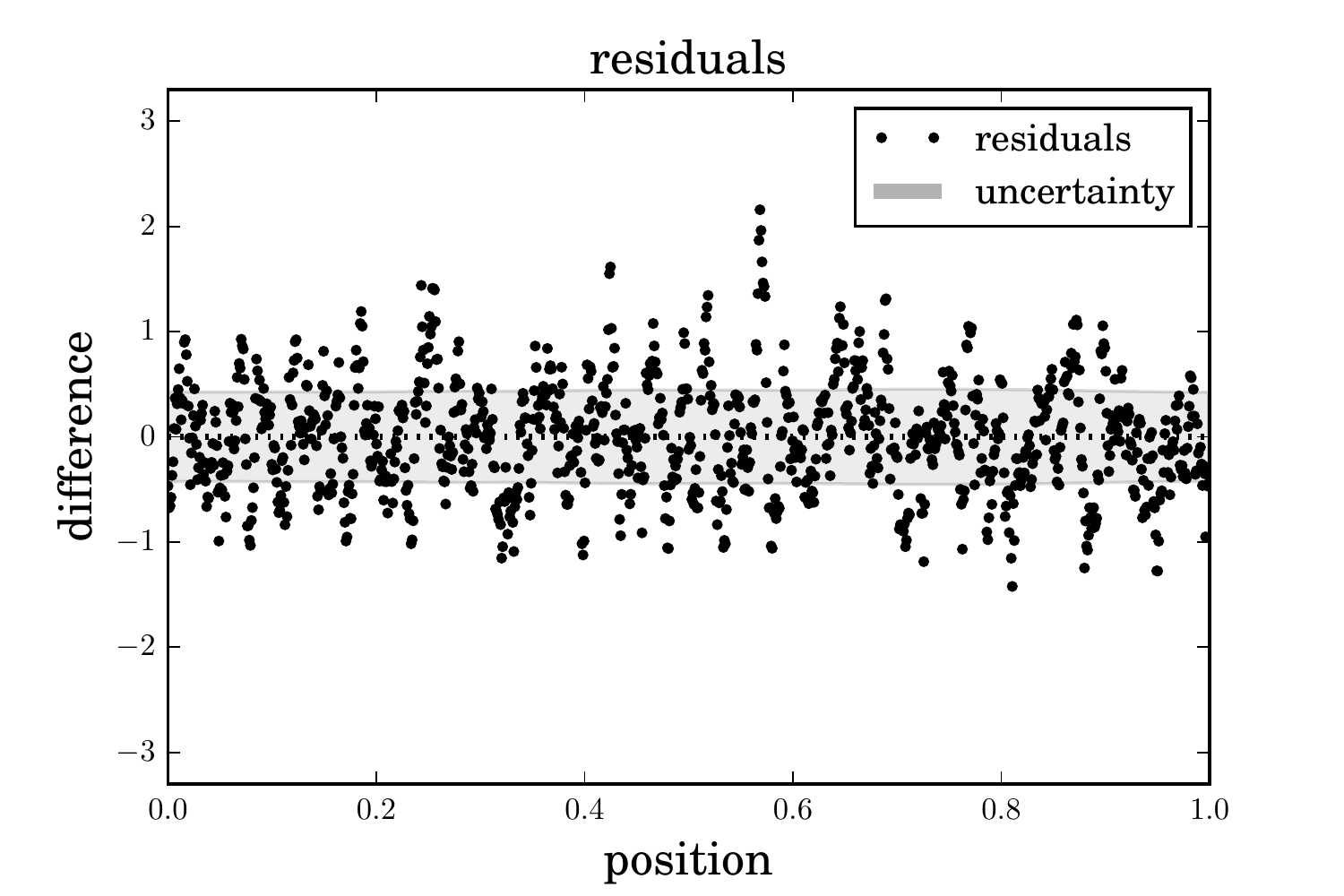}\includegraphics[width=0.5\textwidth]{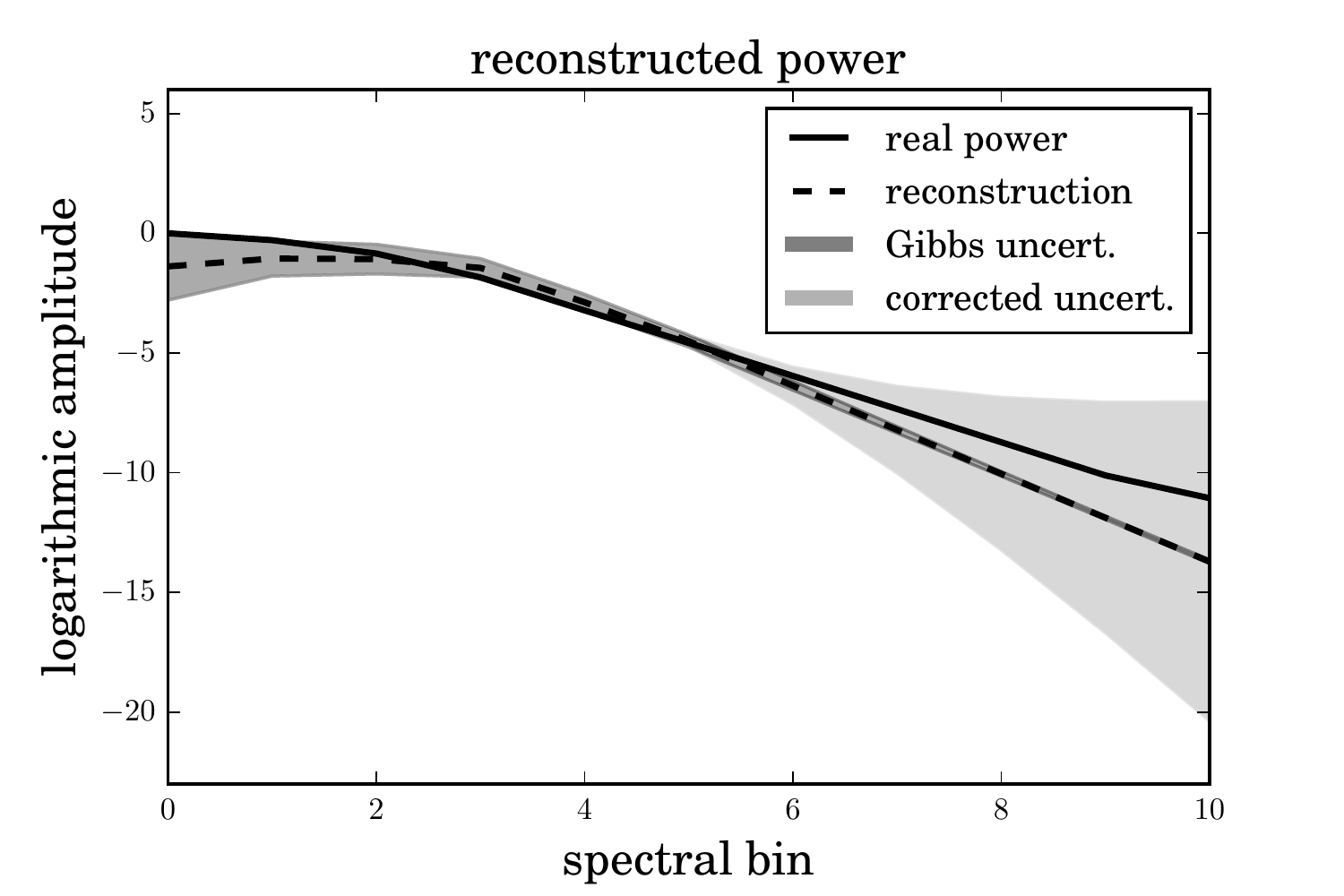}\caption{CSI of a normal field. \emph{Top left:} Normal field $\varphi$ and
noisy data $d$. \emph{Top right:} Reconstructed signal $m$. \emph{Bottom
left:} Reconstruction error $s-m$ and one sigma uncertainty from
$\widehat{D}^{\nicefrac{1}{2}}$. \emph{Bottom right:} Real and reconstructed
log-power spectra $\tau$ and $t$, the latter augmented by its 1-$\sigma$
uncertainty from $\widetilde{\Theta}^{\nicefrac{1}{2}}$ , the uncorrected
uncertainty (narrow, dark gray), and by $\widetilde{\Theta}'^{\nicefrac{1}{2}}$,
the corrected uncertainty (wider, light gray). \label{fig:N-field-rec}}
\end{figure*}

The fix point of the system of equations worked out above exhibits
a defect. Eqs.\ \ref{eq:t_dot} and \ref{eq:t_fix_point} state that
at the fix point we should have
\[
\theta=\alpha-1+\frac{\varrho}{2}+T\,t.
\]
 To simplify matters, let us assume we have an uninformative spectral
hyperprior ($\alpha=1$, $q=0$), standard temperature ($\mathcal{T}=1$),
and the obtained spectrum is power-law like ($T\,t\approx0$). Then
the spectral uncertainty covariance is $\Theta=\left(T+\widetilde{\nicefrac{\varrho}{2}}\right)^{-1}-A_{\Theta}$,
which turns out to be very narrow. First, we note that the term $-A_{\Theta}$
only tightens an already tight variance $F=\left(T+\widetilde{\nicefrac{\varrho}{2}}\right)^{-1}$.
The latter does not depend on the data, as $T$ and $\varrho$ are
completely predefined. Thus, the spectral uncertainty seems to be
nearly independent of how the measurement was performed, except for
a small correction by $-A_{\Theta}$, which only worsens the problem
of a too tight uncertainty covariance. The spectral uncertainty should,
however, depend on the data quality and be larger for less informative
data. Here it seems, that all signal field degrees of freedom provide
confidence to the spectral determination, irrespectively whether they
were directly determined by the prior and data (the term $q+\frac{1}{2r}\,\mathrm{Tr}[\left(m-C\right)\left(m-C\right)^{\dagger}\,\mathbb{P}]$
in $w$) or only guessed by a covariance (the term $\frac{1}{2r}\,\mathrm{Tr}\left[D\,\mathbb{P}\right]$
in $w$). 

In order to fix this, we propose to modify the formula for $F$ to
\begin{eqnarray}
F & = & \mathcal{T}\,\left(T+\widetilde{\theta'}\right)^{-1}\text{ with}\nonumber \\
\theta' & = & \left\{ q+\frac{w'}{2\,r}\right\} \,e^{-t'}\text{ and}\\
w' & = & \mathrm{Tr}\left[\left(m-C\right)\left(m-C\right)^{\dagger}\,\mathbb{P}\right]=w-\mathrm{Tr}\left[D\,\mathbb{P}\right].\nonumber 
\end{eqnarray}
This way, the uncertainty information $D$ is not counted when the
certainty of $\tau$ is estimated. Effectively, this corresponds to
using a reduced number of degrees of freedom $\varrho'<\varrho$ while
estimating the spectral uncertainty. For our simplified case ($\alpha=1$,
$q=0$, $T\,t\approx0$) we have $\nicefrac{\varrho'}{\varrho}=\nicefrac{w'}{w}$.

The deeper reason for this defect delivers also a justification for
our fixing strategy. The Gibbs free energy approach is equivalent
to minimizing the Kullback-Leibler divergence 
\begin{equation}
d_{\mathrm{KL}}(\widetilde{\mathcal{P}}||\mathcal{P})=\int\mathcal{D}s\,\widetilde{\mathcal{P}}(s|d)\,\ln\left[\widetilde{\mathcal{P}}(s|d)/\mathcal{P}(s|d)\right]
\end{equation}
 of our approximated posterior $\widetilde{\mathcal{P}}(s|d)=\mathcal{G}(s-\overline{s},D)$
to the correct one, $\mathcal{P}(s|d)$. This is, however, only an
approximation, and not even the most optimal one. It was shown from
first principles in \citep{2016arXiv161009018L} that the information
optimal scheme would minimize the inverse Kullback-Leibler divergence,
$d_{\mathrm{KL}}(\mathcal{P}||\widetilde{\mathcal{P}})$. It turns
out that this demands that the first and second moments of $\mathcal{P}(s|d)$
are calculated and used to specify the Gaussian approximation $\widetilde{\mathcal{P}}(s|d)$.
The Gibbs free energy approach was, however, just introduced to facilitate
the calculation of these moments, which are otherwise hard to calculate.

Fortunately, the spectral uncertainty $\sigma_{k}^{2}$ of a Gaussian
random field in a spectral band $k$ of which $\varrho_{k}'$ modes
were measured turns out to be $\sigma_{k}^{2}=\nicefrac{2}{\varrho_{k}'}$
in the absence of any prior information ($\alpha=1$, $q=0$, $T=0$).
The fix proposed above just introduces a correction so that only the
number of effectively measured spectral degrees of freedom, $\varrho'$,
and not the total number, $\varrho$, are used in the calculation
of spectral uncertainty such that in the prior free case the well
known result $\sigma_{k}^{2}=\nicefrac{2}{\varrho_{k}'}$ is recovered. 

\subsection{Numerical steps}

\begin{figure*}[t]
\includegraphics[bb=50bp 40bp 402bp 288bp,clip,width=0.5\textwidth]{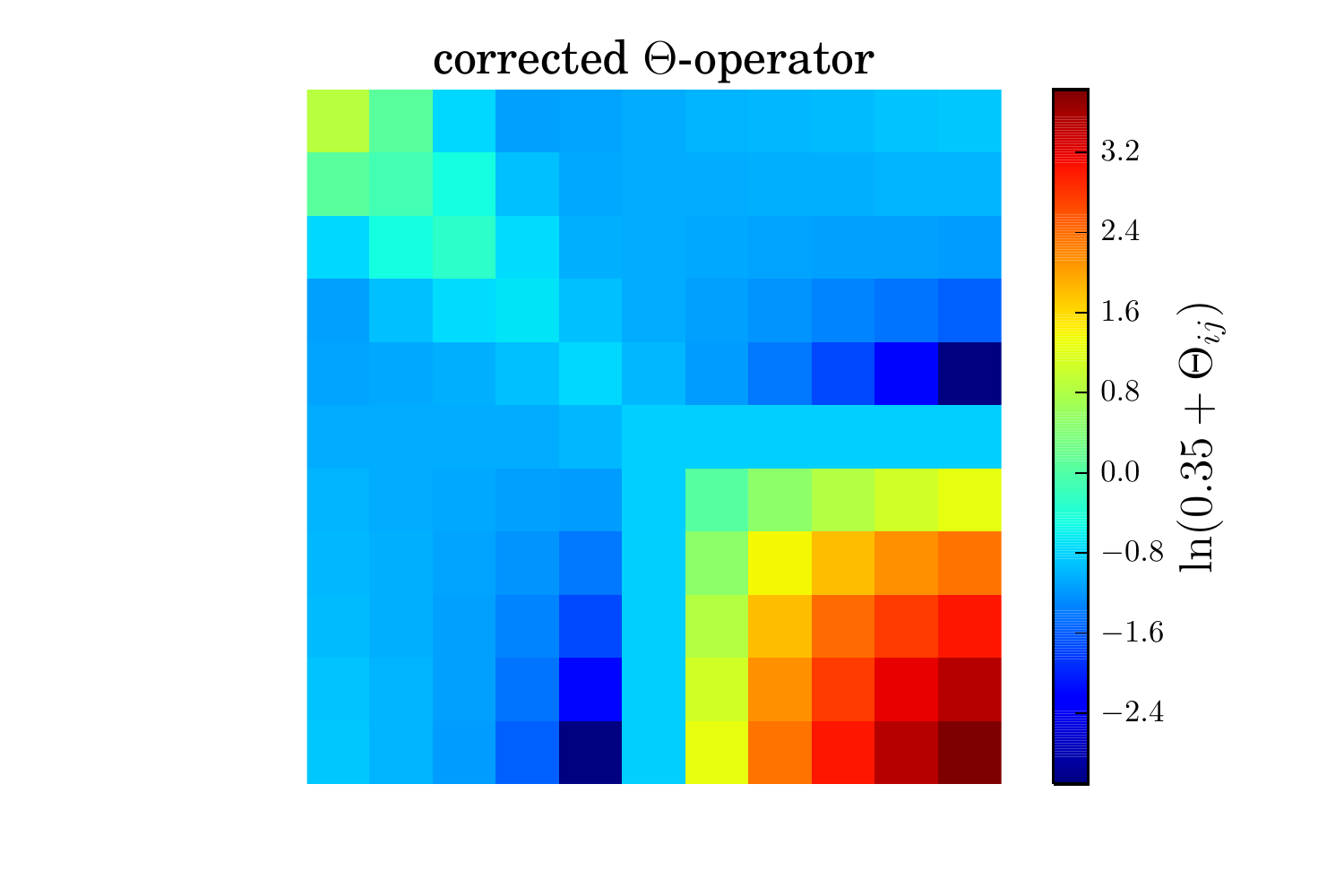}\includegraphics[width=0.5\textwidth]{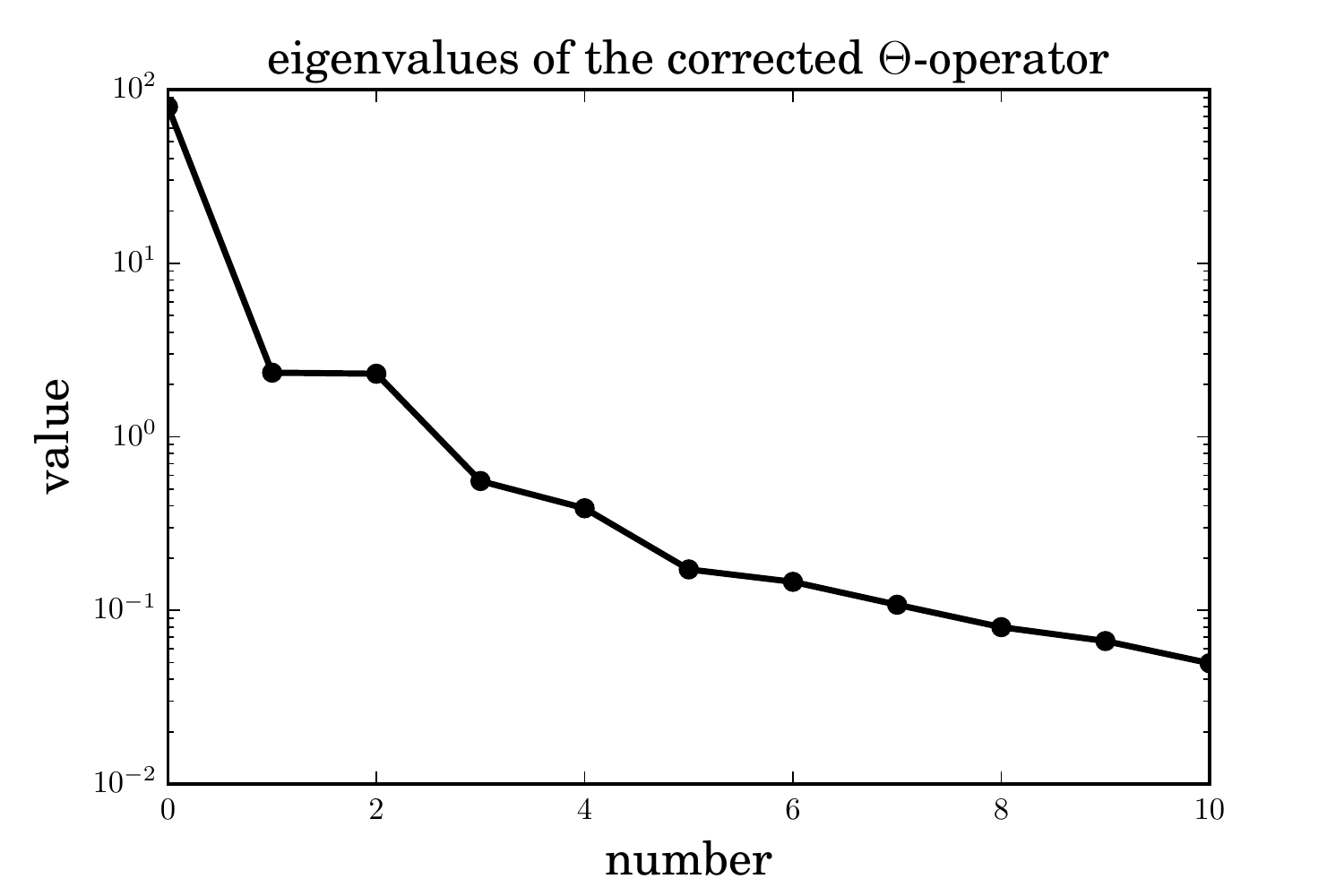}

\includegraphics[width=0.5\textwidth]{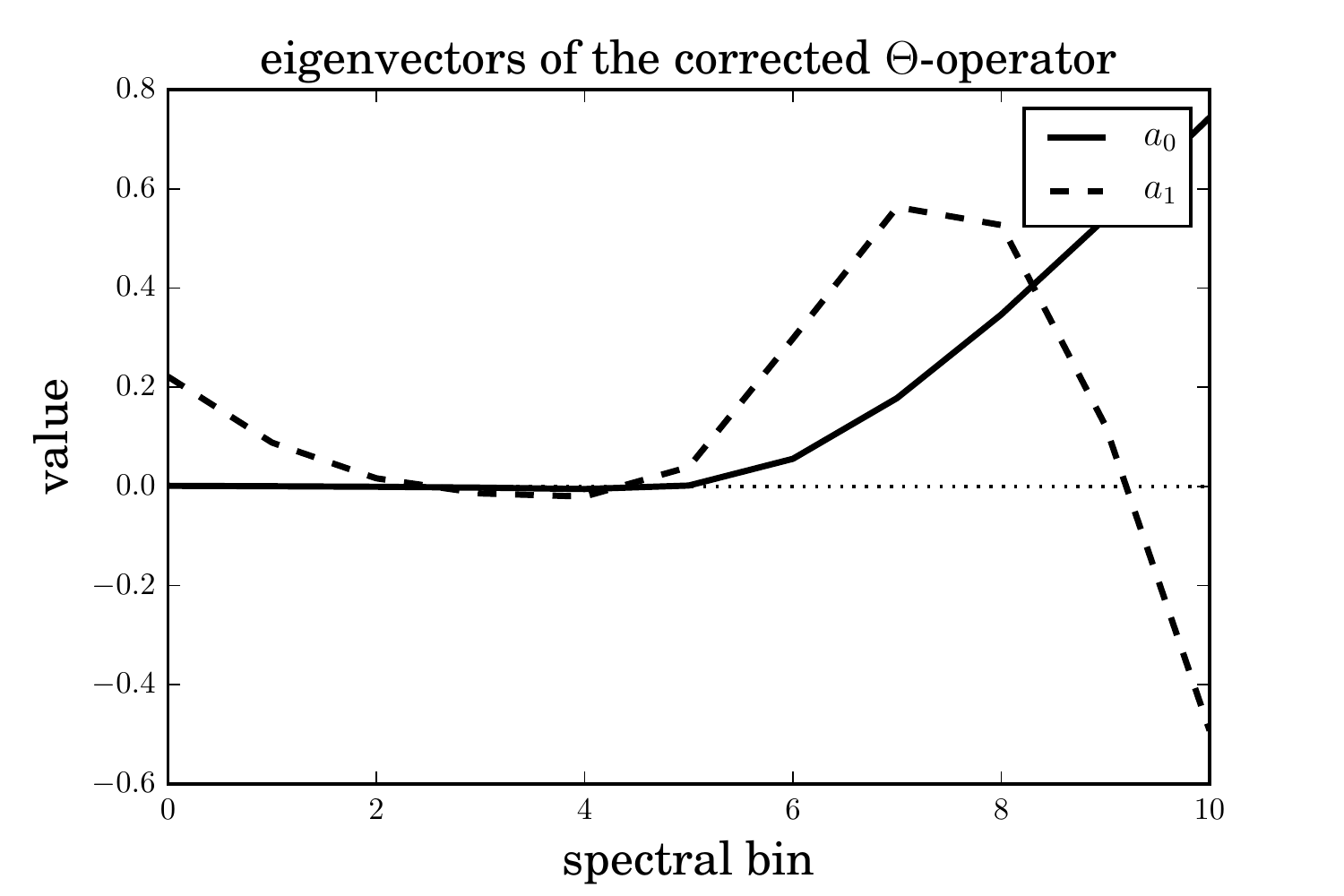}\includegraphics[width=0.5\textwidth]{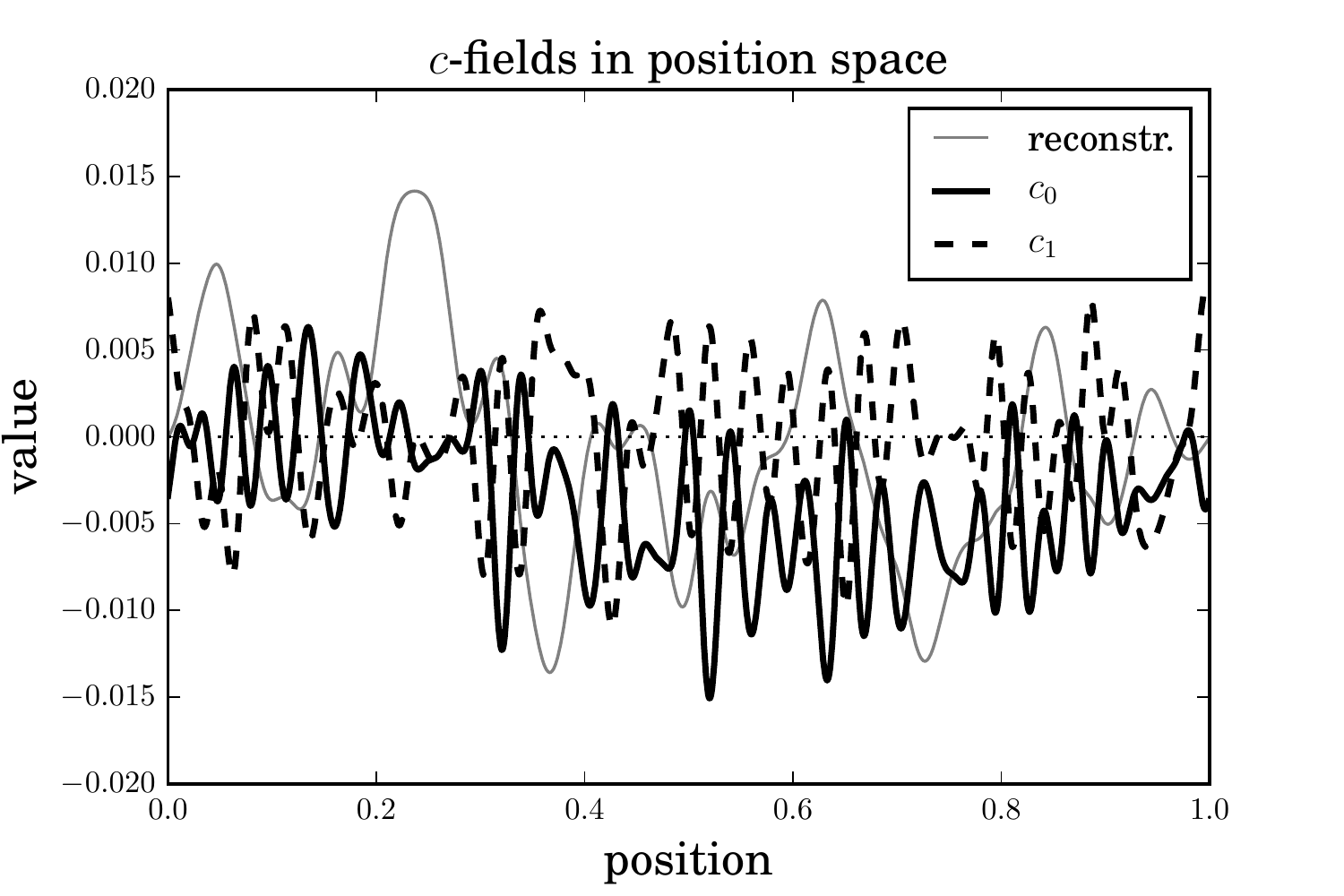}

\caption{\emph{Top Left:} Corrected spectral uncertainty matrix $\Theta'$
for the normal field. The large scale modes are in the top left corner.
Note the anti-correlations (dark blue).\emph{ Top right:} Eigenspectrum
of $\Theta$'. \emph{Bottom left:} Largest two eigenvectors of $\Theta'$.
\emph{Bottom right:} Corresponding $c$ fields of the spectral-spatial
uncertainty cross correlation $C$, superimposed on re-scaled field
reconstruction $m$ (gray). \label{fig:Spectral-Uncertainty-Structure}}
\end{figure*}

\subsubsection{Notation}

Let $\boldsymbol{z}=(m\,,t\,,\,c,\,D,\Theta)$ be the argument vector
of the Gibbs free energy $G(\boldsymbol{z})$ we intend to minimize,
or, alternatively, the vector of the fields we track dynamically,
$\boldsymbol{z}=(m,\,t,\,c,\,A_{c},\,A_{\Theta},\,\widehat{D},\,\widetilde{\Theta},\ldots)$.
We split $\boldsymbol{z}=(\boldsymbol{x},\,\boldsymbol{y})$ into
the part for which we follow explicitly the gradient dynamics, $\boldsymbol{x}=(m,t\,)=\overline{s}$,
and the part evolving implicitly or estimated stochastically $\boldsymbol{y}=(c,\,A_{c},A_{\Theta},\widehat{D}\,,\widetilde{\Theta},\ldots)$.

The \textsc{FrEE} dynamics for $\boldsymbol{x}$ is driven by its
Gibbs free energy force 
\begin{equation}
\boldsymbol{f}=-\frac{\partial G(\boldsymbol{\boldsymbol{x,\,\boldsymbol{y}}})}{\partial\boldsymbol{x}},
\end{equation}
which imposes on $\boldsymbol{x}$ the velocity 
\begin{equation}
\frac{d\boldsymbol{x}}{d\mathtt{t}}=\boldsymbol{\Gamma\,}\boldsymbol{f}.\label{eq:X_dyn}
\end{equation}
Here, $\boldsymbol{\Gamma}$ is a positive definite, inverse friction
coefficient matrix, which fixes the different units of energy ($G$),
time $(\mbox{\ensuremath{\mathtt{t}}})$, and space ($\boldsymbol{x}$)
in this equation, which will be specified next. 

\subsubsection{Steering }

The operator $\Gamma$ steers the algorithm. We get a Newton scheme
if $\boldsymbol{\Gamma}=S=(\delta^{2}G/\delta x\,\delta x^{\dagger})^{-1}$
and the time steps are $\delta\mathtt{t}=1$, such that
\begin{equation}
\boldsymbol{x}\leftarrow\boldsymbol{x}+\boldsymbol{\Gamma}\,\boldsymbol{f}.
\end{equation}
In case of non-convex problems, S can develop temporarily negative
eigenvalues, which would spoil the numerical scheme. 

To avoid this problem, we propose to steer the degrees of freedom
of $\boldsymbol{x}$ individually by adding a mass matrix-like term,
\begin{align}
\boldsymbol{\Gamma} & =\left(S^{-1}+\boldsymbol{\widehat{\eta}}\right)^{-1},\label{eq:gamma_final-1}
\end{align}
with a suitably chosen damping vector $\boldsymbol{\eta}$ with $\boldsymbol{\eta}>0$
in all components. $\boldsymbol{\eta}=\lambda$ with a global tuning
parameter $\lambda$ would correspond to Levenberg's algorithm \citep{citeulike:10796881}
and $\boldsymbol{\widehat{\eta}}=\lambda\boldsymbol{f}\,\boldsymbol{f}^{\dagger}$
would to the Levenberg\textendash Marquardt algorithm \citep{doi:10.1137/0111030}.
Here, we pragmatically propose to tune the components of $\boldsymbol{\eta}$
on the fly.  

To this end, we set $\boldsymbol{\eta}=\widehat{S^{-1}}\,\boldsymbol{\eta'}$
to introduce with $\widehat{S^{-1}}=(\widehat{E^{-1}},\widetilde{F^{-1}})^{\dagger}$
typical scales, so that $\boldsymbol{\eta'}$ is dimensionless. The
latter is component wise steered via 
\begin{equation}
\boldsymbol{\eta'}\leftarrow\boldsymbol{\eta'}\begin{cases}
f_{\eta',\text{ acc}} & \delta\boldsymbol{x}_{n}\delta\boldsymbol{x}_{n-1}>0\\
f_{\eta',\text{ break}} & \delta\boldsymbol{x}_{n}\delta\boldsymbol{x}_{n-1}\le0
\end{cases},
\end{equation}
where $\delta\boldsymbol{x}_{n}$ is the update vector of step $n$
and $f_{\eta',\text{ acc}}=0.5$ and $f_{\eta',\text{ break}}=3$
are our empirical choices for the numerical tuning parameters. $\boldsymbol{\eta'}\ge0.3$
is ensured in every step, to keep the scheme ready to damp any degree
of freedom that starts to get unstable. 

This way, any oscillating parameter gets damping. When the scheme
reached the minimum, all parameters will oscillate around the equilibrium
position and $\delta\boldsymbol{x}_{n}\delta\boldsymbol{x}_{n-1}$
will have random signs. 
\begin{figure}[t]
\includegraphics[bb=100bp 40bp 325bp 278bp,clip,height=0.22\textwidth]{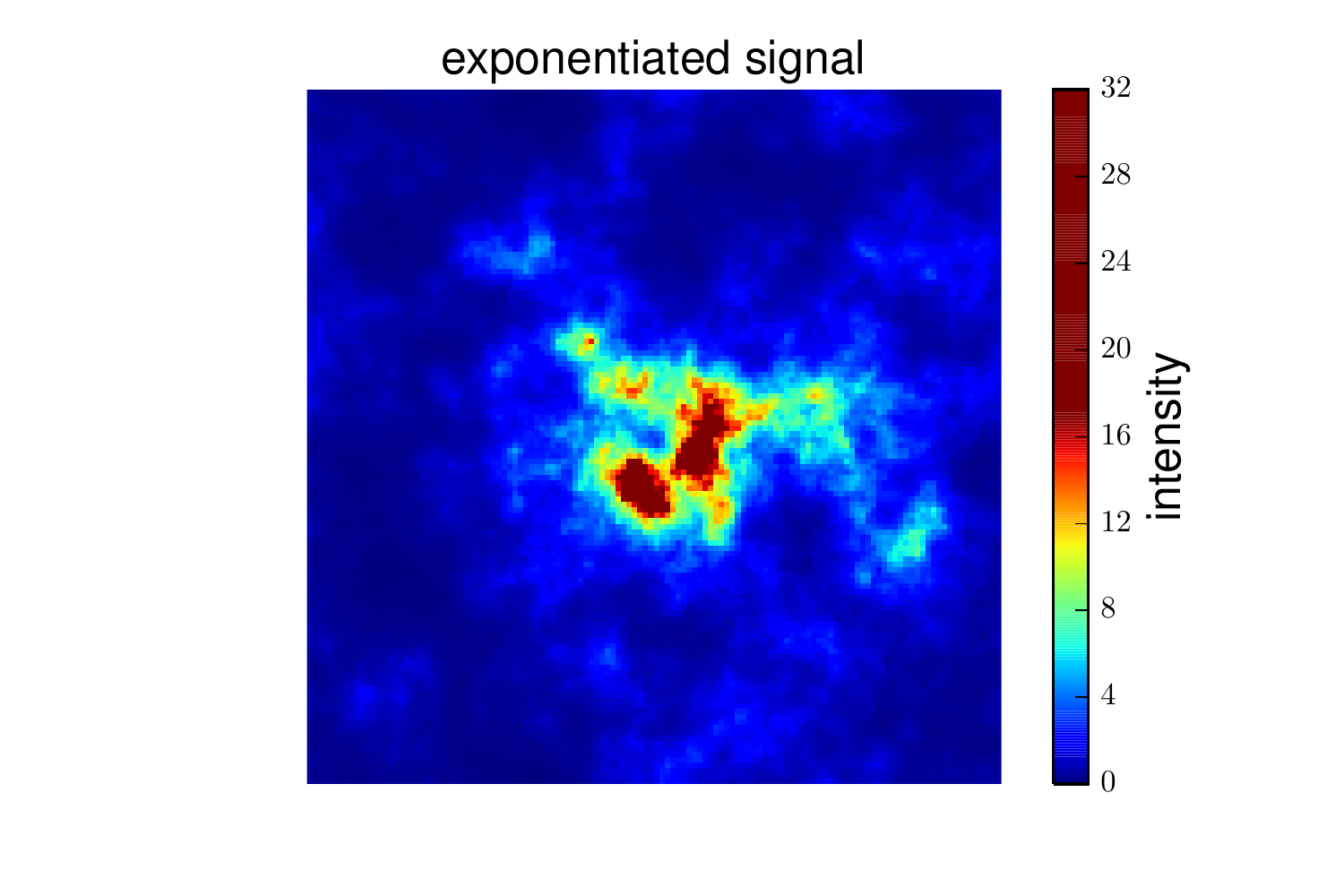}\includegraphics[bb=100bp 40bp 385bp 278bp,clip,height=0.22\textwidth]{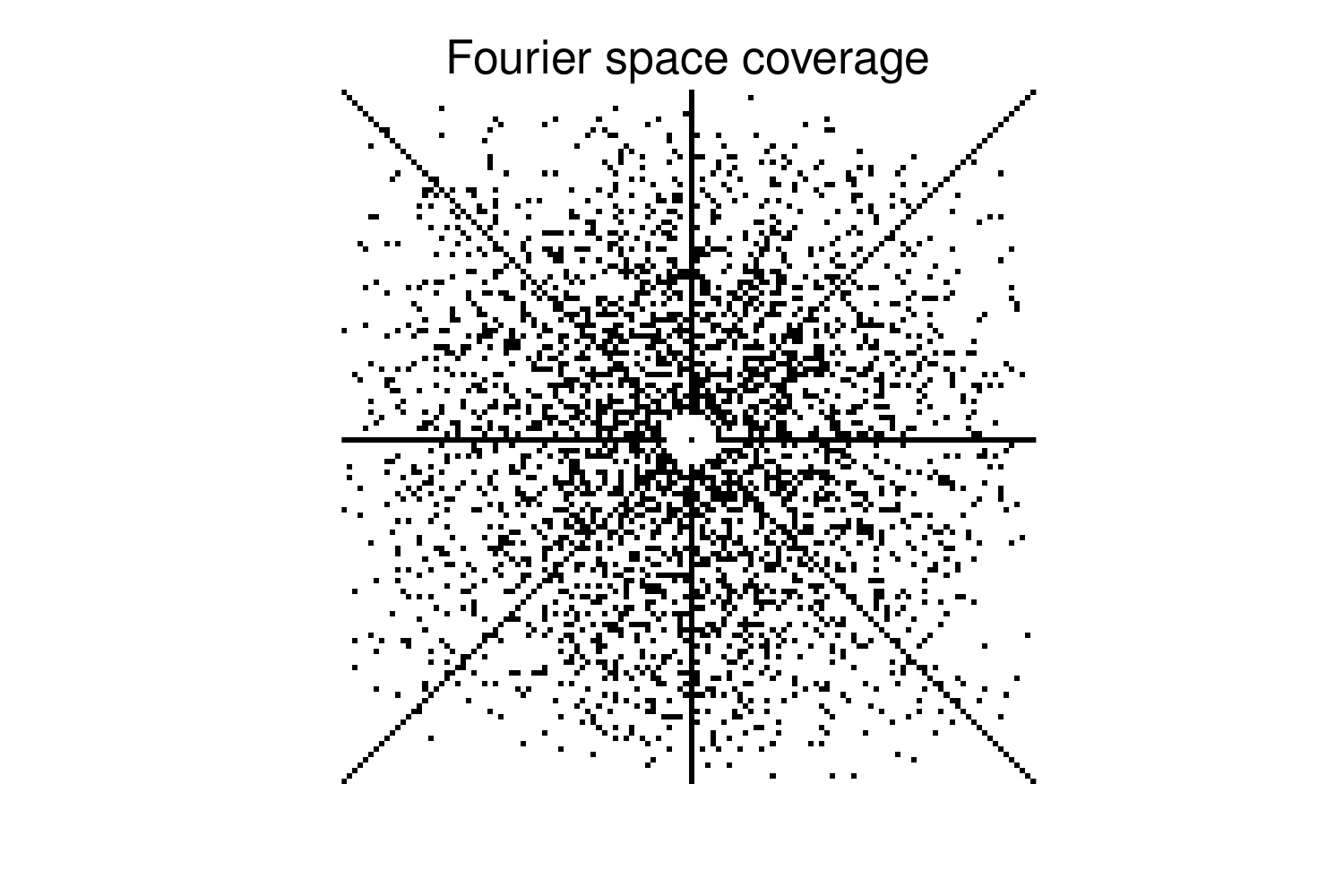}

\includegraphics[bb=100bp 40bp 325bp 278bp,clip,height=0.22\textwidth]{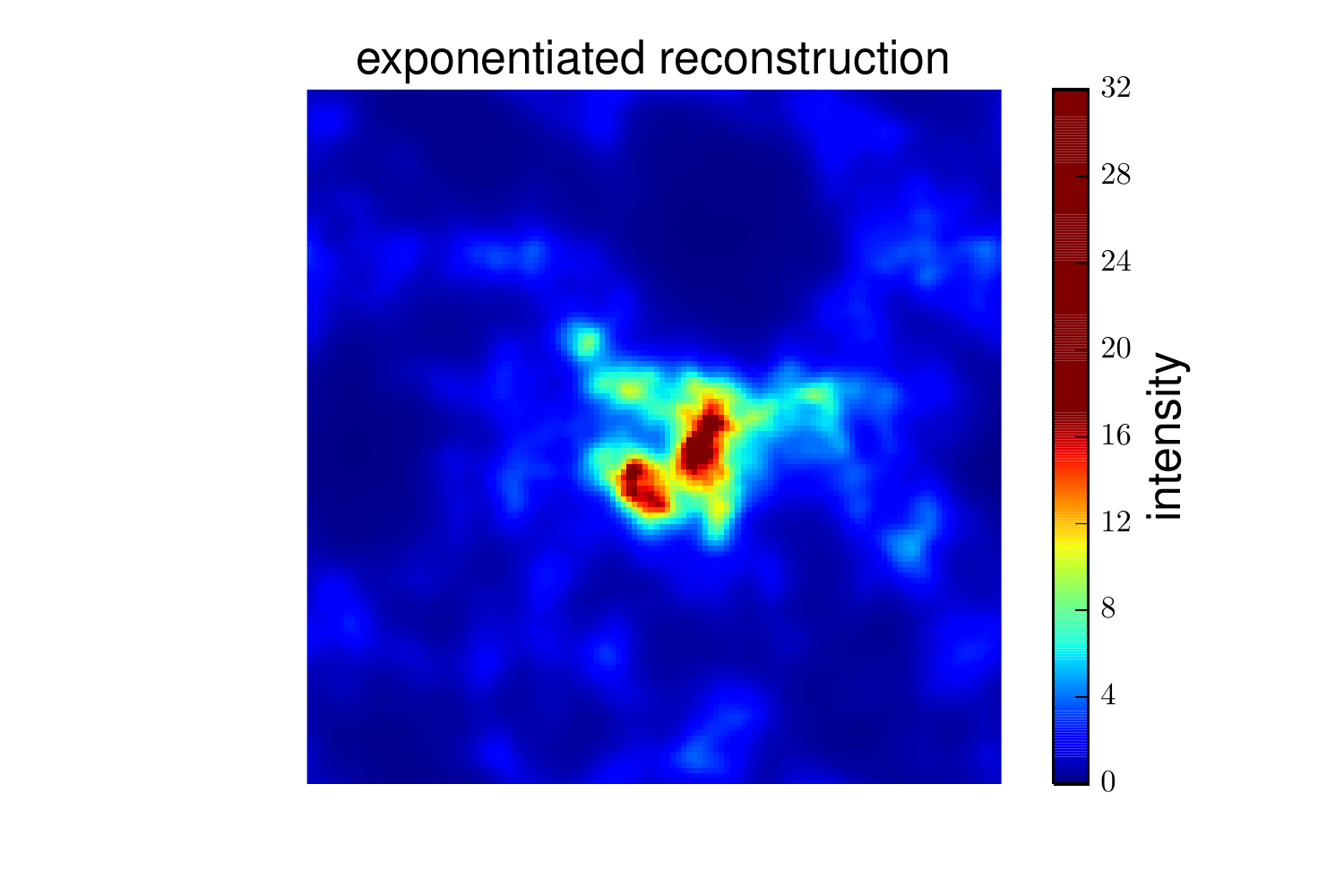}\includegraphics[bb=100bp 40bp 385bp 278bp,clip,height=0.22\textwidth]{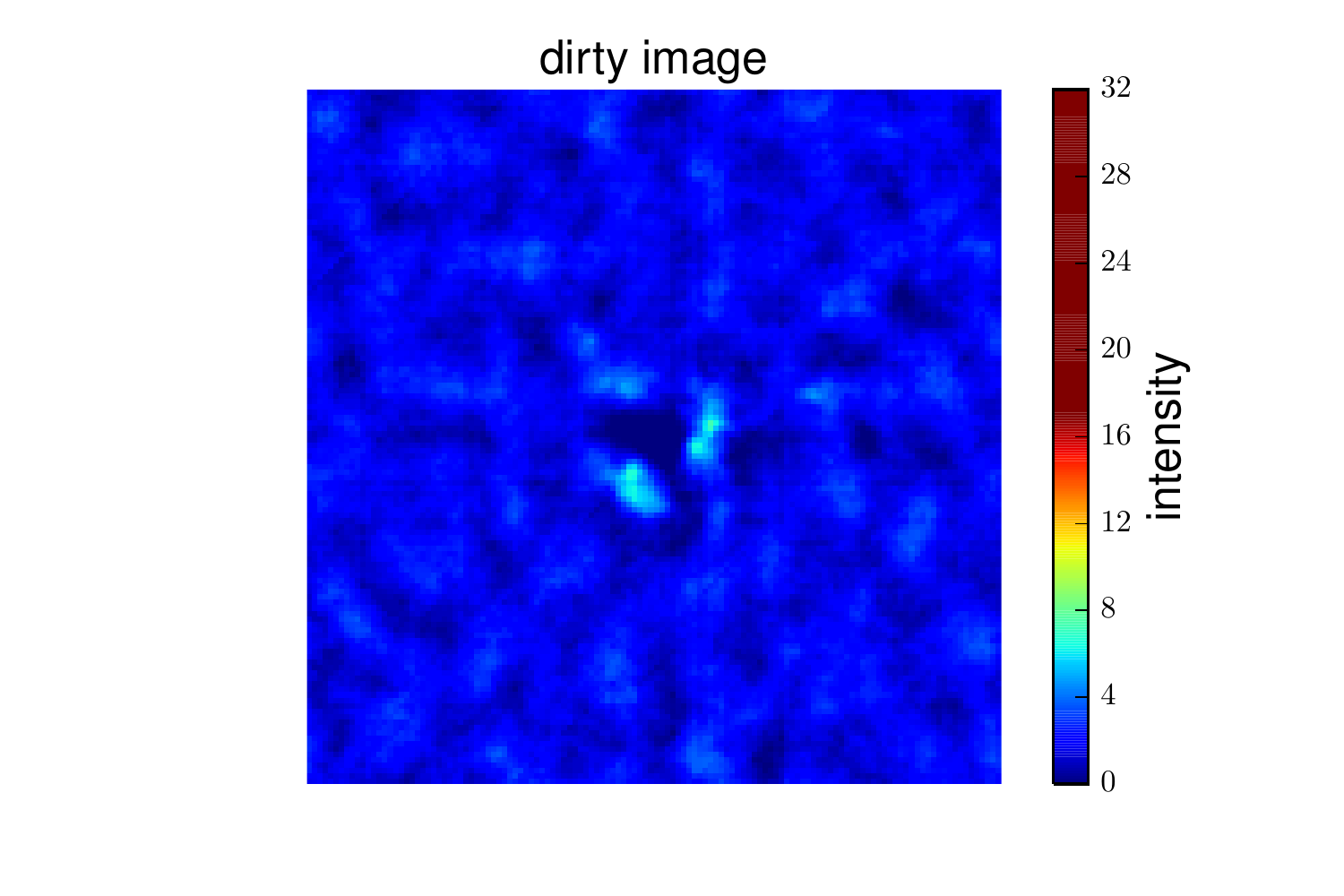}

\caption{\emph{Top Left:} Sky brightness as a realization of a LN field\emph{
Top right:} Fourier-space coverage of the simulated interferometer.
\emph{Bottom left:} Reconstruction. \emph{Bottom right:} ``Dirty
image'', the back projected data.\label{fig:LogNormalPics}}
\end{figure}

\subsubsection{Implementation simplifications }

Not all the discussed terms are essential in every situation. The
related critical filter scheme for CSI lacks many of the here discussed
terms, but provides acceptable results. Therefore, a number of simplifications
are adapted here to reduce the computational complexity:
\begin{itemize}
\item The terms $A_{\Theta}$ and $A_{D}$ are approximated as in Eq.\ \ref{eq:A_D_approx}.
This truncates the mutual dependence of the $D$ and $\Theta$ operators.
\item The field-to-spectrum uncertainty cross-correlation term $C$ is set
to zero in most numerical runs, as no significant effect of it on
the numerical speed, or the results could be detected. An inspection
of $C$ in Sect.\ \ref{subsec:Field-to-spectrum-cross-correlat}
reveals that it is of a low relative amplitude in the test cases we
investigate.
\item Steering was only applied to the $t$-dimension, the spectral components
of $\boldsymbol{x}$. In case of very stiff problems, an extension
to $m$ might be necessary. This will be used for the LN problem presented
below, which is stiff due to a response function mimicking a radio
interferometer.
\item The CG used to calculate steps of the Newton scheme was run with moderate
accuracy, as all steps are only intermediate. Experiments with low
accuracy CG had a worse performance. 
\end{itemize}
The implementation of the CSI algorithm was done in NIFTy 1.0 \citep{2013ascl.soft02013S}.
\begin{figure}[t]
\includegraphics[bb=0bp 0bp 432bp 288bp,clip,width=0.5\textwidth]{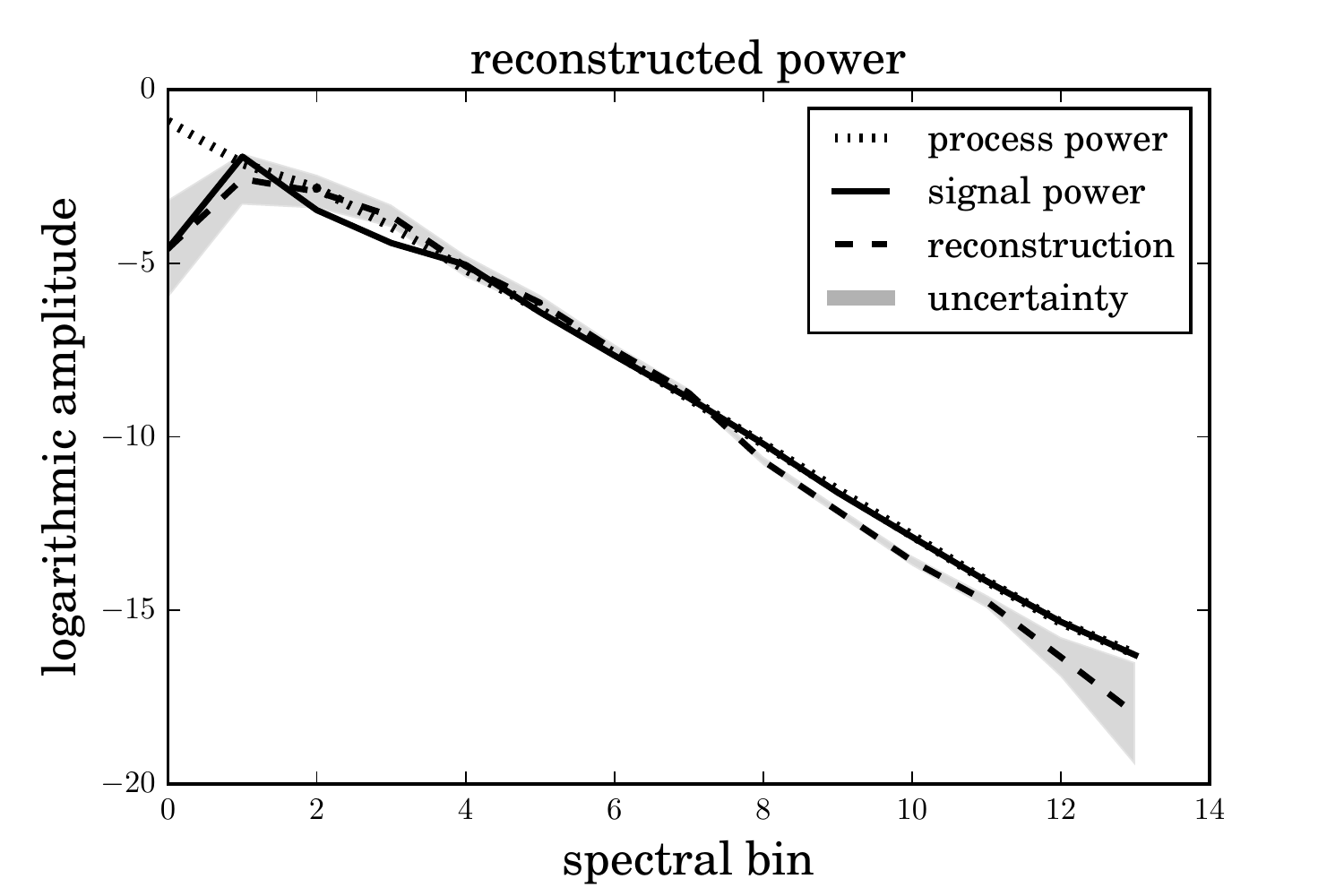}

\caption{Power spectra of the signal realization shown in Fig.\ \ref{fig:LogNormalPics}
(solid line) and of the statistical process, which generated it (dotted).
The spectrum reconstruction (dashed) including its uncertainty (gray)
is shown as well. \label{fig:NL_spectrum}}
\end{figure}

\begin{figure*}[t]
\includegraphics[bb=0bp 0bp 432bp 288bp,clip,width=0.5\textwidth]{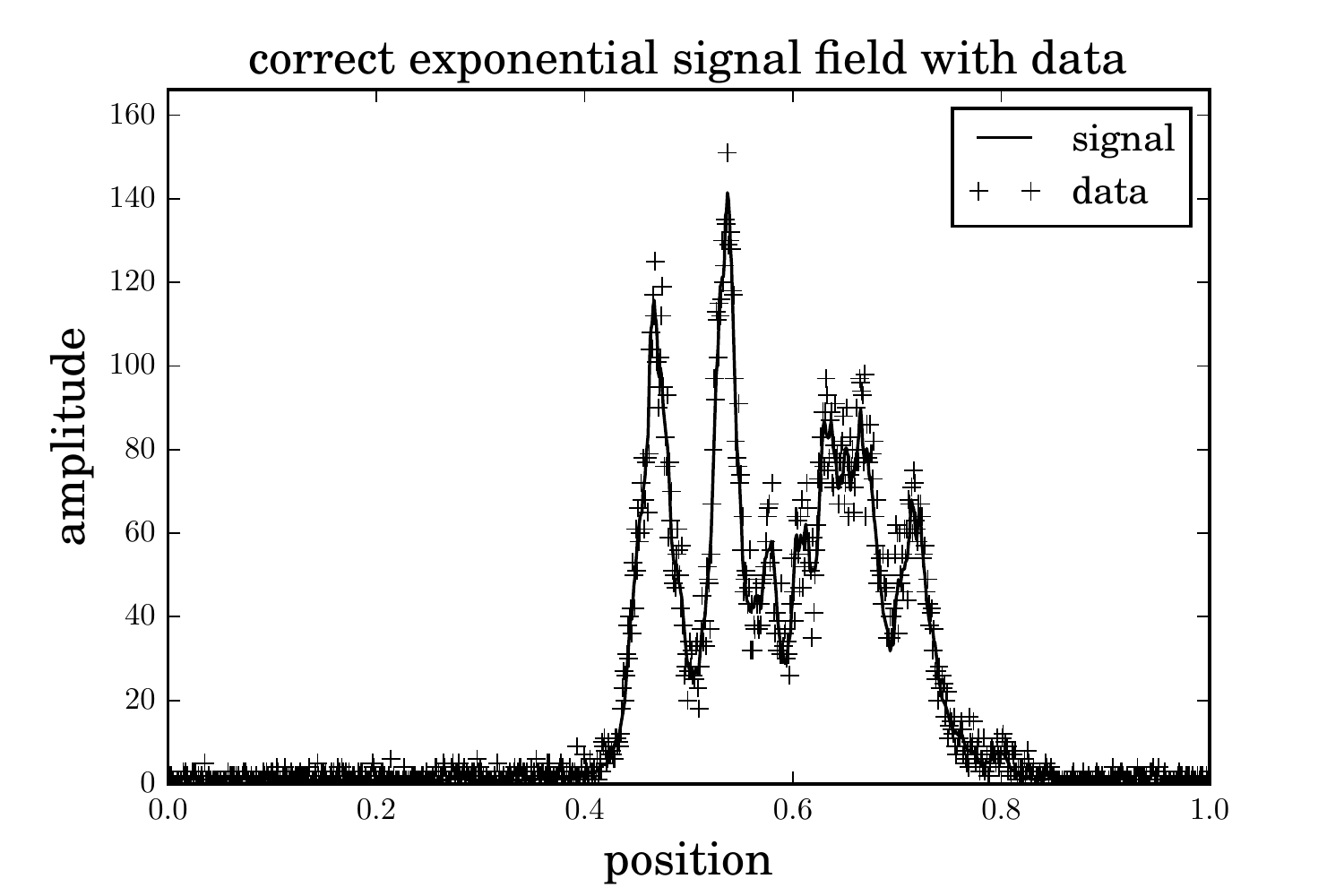}\includegraphics[width=0.5\textwidth]{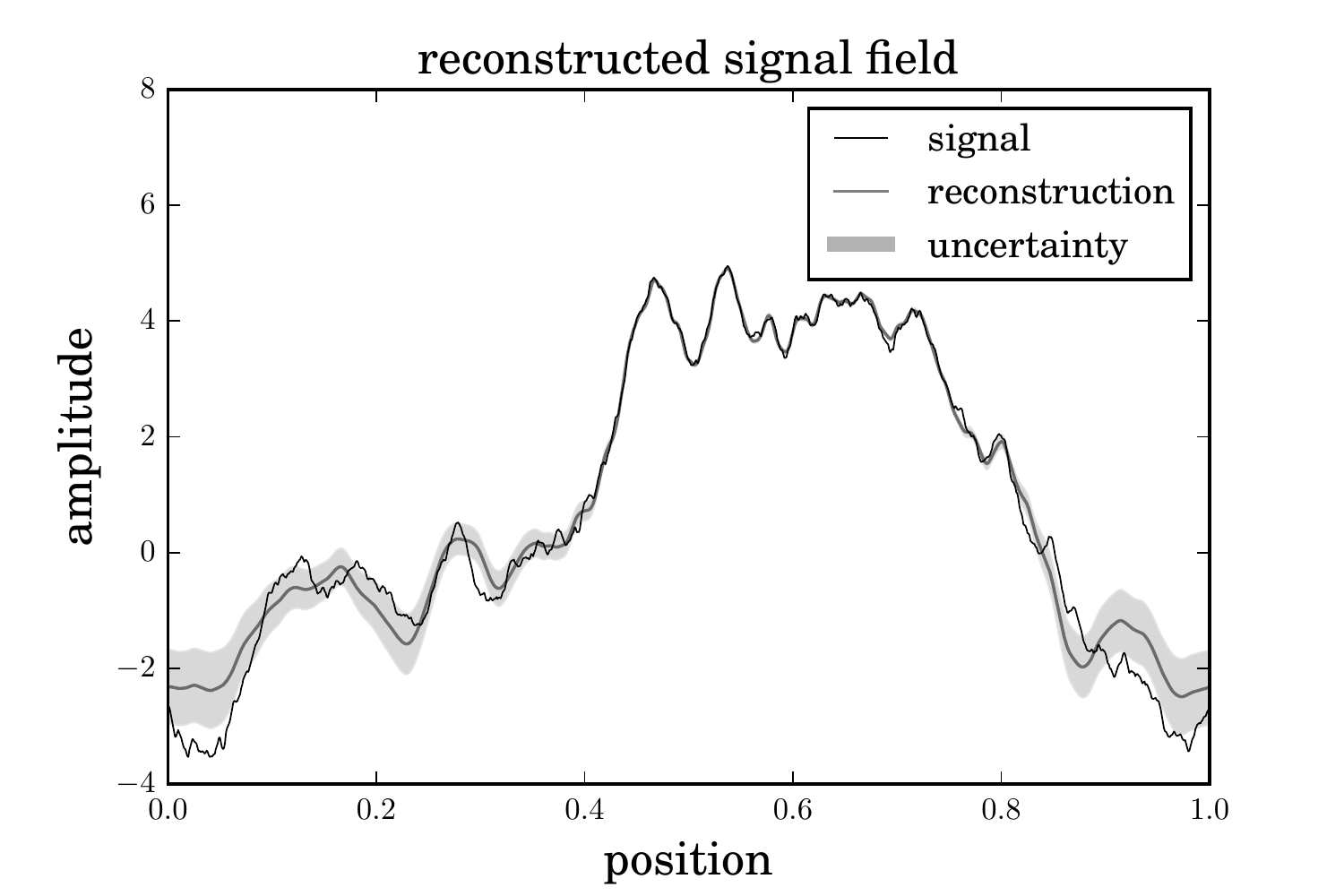}

\includegraphics[width=0.5\textwidth]{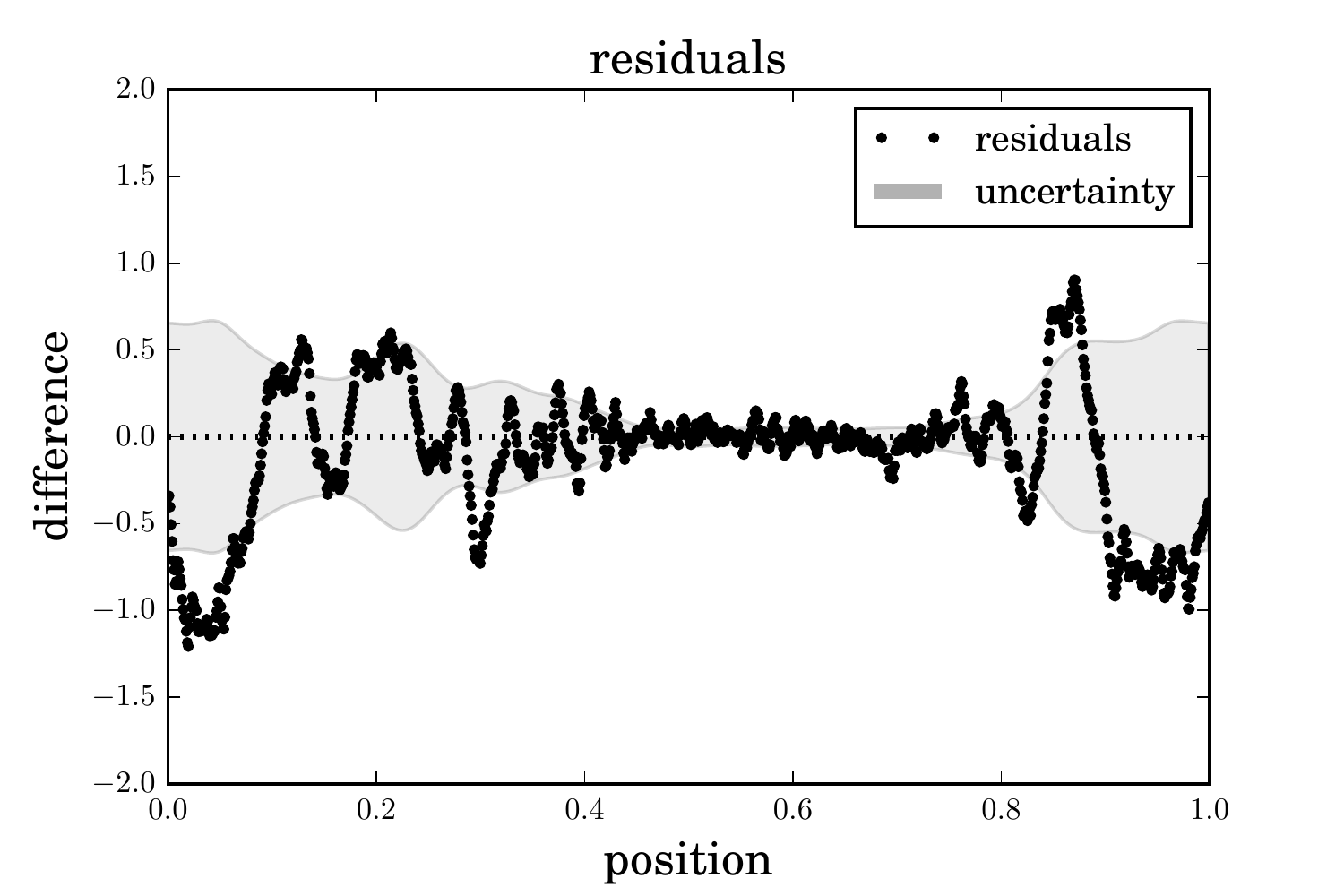}\includegraphics[width=0.5\textwidth]{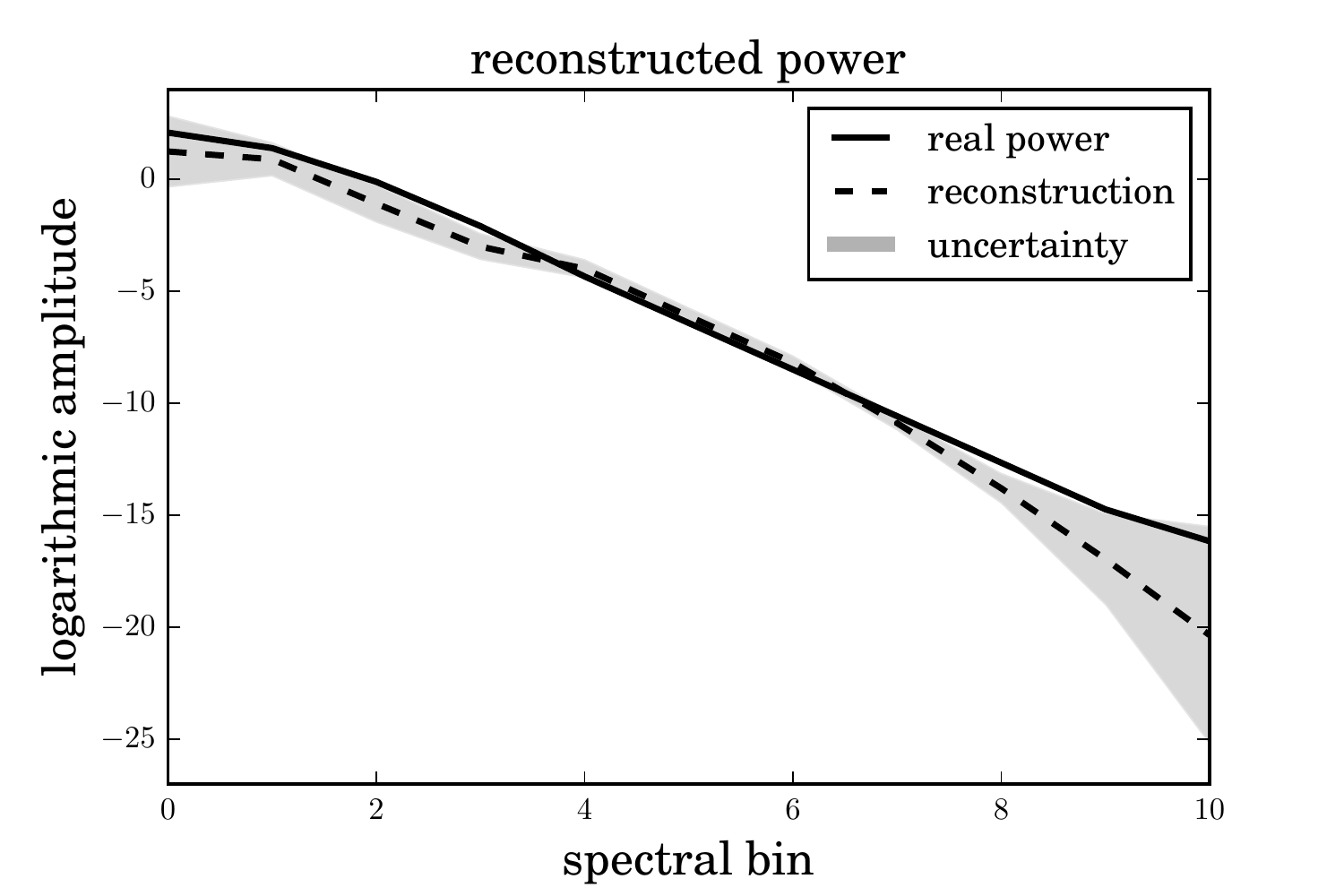}

\caption{PLN field reconstruction problem.\emph{ Top Left:} Exponentiated signal
$e^{\varphi}$ and Poisson data drawn from this. \emph{Top right:}
Signal field $\varphi$and its reconstruction $m$ including the remaining
uncertainty. Note the higher fidelity in the high count regions. \emph{Bottom
left:} Reconstruction errors $\varphi-m$. \emph{Bottom right:} Correct
and reconstructed power spectrum. \label{fig:PLN}}
\end{figure*}

\section{Examples\label{sec:Examples}}

To illustrate the performance for normal (N), log-normal (LN), and
Poisson-log-normal (PLN) fields, the CSI algorithm was applied to
mock data generated with such statistics. We report about the results.

\subsection{Normal field}

\subsubsection{Spatial and spectral reconstruction}

A normal field, the power spectrum of its generating statistical process,
noisy data tracing a Gauss-convolved version of the field, and the
reconstruction of the field and its spectrum are displayed in Fig.\ \ref{fig:N-field-rec}.
The field is well reconstructed within the uncertainties, which are
much smaller than the scatter of the data. The power spectrum, however,
is reconstructed with an overconfidence in the higher spectral bins
(darkly shaded in the bottom right panel of Fig.\ \ref{fig:N-field-rec})
unless the fix to the effectively measured degrees of freedom as described
in Sect.\ \ref{subsec:Fixing-the-fix} is applied (lightly shaded).
It is very apparent that this correction is necessary as the Gaussian
point spread function has erased a large part of the information of
the signal field spectrum on small spatial scales. We will adapt this
correction for the remaining discussion of the N example and also
for the LN and PLN examples.

\subsubsection{Spectral uncertainty structure\label{subsec:Field-to-spectrum-cross-correlat}}

The CSI algorithm provides the full spectral uncertainty covariance
matrix, which in its corrected form (see discussion in Sect.\ \ref{subsec:Fixing-the-fix})
is displayed for our example in the top left panel of Fig.\ \ref{fig:Spectral-Uncertainty-Structure}.
The large uncertainty of the highest Fourier band powers is obviously
correlated, as the spectral smoothness prior prevents these from varying
independently. The eigenvalues of this uncertainty covariance matrix
span two orders of magnitude (top right panel). The eigenvector $a_{0}$
corresponding to the largest eigenvalue (see bottom left panel) expresses
the substantial and correlated uncertainty of the small-scale power
spectrum (see bottom right corner of top left panel). The next important
eigenvector $a_{1}$ allows for structure in this part of the spectrum
via a sign change and captures the main spectral uncertainty for large
spatial scales (see top left corner of top left panel). 

Finally, Fig.\ \ref{fig:Spectral-Uncertainty-Structure} displays
the corresponding $c$-fields, that were calculated according to Eq.\ \eqref{eq:c_i},
and which form, together with their eigenvectors, the spectral-spatial
uncertainty covariance operator $C$ according to $C=\sum_{i}c_{i}a_{i}^{\dagger}$
(Eq.\ \eqref{eq:C_representation}). The $c_{i}$-field expresses
how the reconstruction $m$ changes with a change of the spectrum
according to the corresponding spectral uncertainty eigenvector $a_{i}$
and vice versa. As the $c$-fields are filtered versions of the reconstruction
$m$, a scaled version of this is displayed in Fig.\ \ref{fig:Spectral-Uncertainty-Structure}
for comparison. 

\subsection{Log-normal field}

A LN field is a good representation of diffuse emission on the celestrial
sphere, as it is strictly positive, spatially correlated, and varies
over orders of magnitude in intensity. To present the performance
of the CSI algorithm for LN fields with non-trivial measurement response,
we adapt an idealized radio interferometric measurement situation.
The brightness field (top left panel of Fig.\ \eqref{fig:LogNormalPics})
is measured in its Fourier representation at the Fourier locations
indicated (top right panel). This is a very incomplete coverage, with
about 80\% of the necessary image information missing. Consequently,
the back projected data (bottom right panel), the so called \emph{dirty
image} in radio astronomical language, is a relative poor representation
of the true sky. In particular it contains areas with nonphysically
negative flux densities. The CSI reconstruction (bottom left) managed
to recover all brighter structures well, and a good fraction of the
dim emission regions. 

The simultaneously recovered power spectrum is shown in Fig.\ \eqref{fig:NL_spectrum}.
Although the power spectrum uncertainty correction of Sect.\ \eqref{subsec:Fixing-the-fix}
is applied, the true and reconstructed spectra deviate by a few sigma
for some of the intermediate spectral bands. The origin of this discrepancy
is not completely clear, as it could be due to the neglected $C$
terms, just a statistical fluctuation, an expression of the strong
non-Gaussianity of the underlying probabilities, or \textendash{}
most likely \textendash{} a combination of such factors.

\subsection{Poisson Log-Normal Field}

Finally, the performance of CSI on PLN reconstruction problems should
be shown. Such problems occur for example in astronomical imaging
with photon counts. Fig.\ \eqref{fig:PLN} shows a PLN field, data
drawn from it, and the reconstruction of the field and its spectrum.
The real signal and spectrum lie well within the corresponding uncertainties.

\section{Conclusion\label{sec:Conclusion}}

Correlated signal inference (CSI) poses numerical challenges. We address
these by the Free Energy Exploration (\textsc{FrEE)} strategy. This
proposes to simplify the inference problem via the construction of
a Gaussian approximation of the posterior, to reconstruct nuisance
and hyper-prior quantities explicitly, and to use stochastic probing
of expensive operator properties, as well as to automatically steer
the step size of the individual signal and power spectrum components.
The resulting \textsc{NIFTy} implementation of CSI accurately reconstructed
normal, log-normal, and Poisson log-normal fields as well as the power
spectra of their generative processes. Uncertainty information on
all these quantities is provided alongside. CSI is therefore mature
for the usage on real data applications, since prototypes of a radio
interferometric and a photon count imaging algorithm performed well.
Furthermore, the \textsc{FrEE} strategy seems also to be sufficiently
robust to be applied to other signal inference problems in information
field theory. 
\begin{acknowledgments}
We acknowledge valuable discussions and comments on the manuscript
by Maksim Greiner, Henrik Junklewitz, Reimar Leike, Natalia Porqueres,
Daniel Pumpe, Theo Steininger. This research has been partly supported
by the DFG Research Unit 1254 and has made use of NASA\textquoteright s
Astrophysics Data System. 
\end{acknowledgments}

\appendix

\section{Log-normal model\label{sec:Log-normal-model}}

\subsection{Field measurement}

Many measured signal fields are strictly positive and vary rather
in magnitude than on a linear scale. An example for such is the sky
brightness $s$. Here a log-normal model is more appropriate where
a Gaussian field $\varphi=\ln s$ is exponentiated before linearly
measured. Thus, we have
\begin{equation}
d=R\,e^{\varphi}+n
\end{equation}
and consequently 
\begin{equation}
\mathcal{P}(d|\varphi,\,\mathrm{LN})=\mathcal{G}(d-R\,e^{\varphi},\,N)
\end{equation}
as well as
\begin{equation}
\mathcal{H}(d|\varphi,\,\mathrm{LN})\widehat{=}\frac{1}{2}\left(d-R\,e^{\varphi}\right)^{\dagger}N^{-1}\left(d-R\,e^{\varphi}\right).
\end{equation}
In case of known field covariance $\Phi$, the joint Hamiltonian
\begin{align}
\mathcal{H}(d,\,\varphi|\Phi,\,\mathrm{LN})= & \mathcal{\,H}(d|\varphi,\,\mathrm{LN})+\mathcal{H}(\varphi|\Phi)\nonumber \\
\widehat{=} & \,\frac{1}{2}\varphi^{\dagger}\Phi^{-1}\varphi\label{eq:LN}\\
 & +\frac{1}{2}\left(d-R\,e^{\varphi}\right)^{\dagger}N^{-1}\left(d-R\,e^{\varphi}\right).\nonumber 
\end{align}
does not permit a linear estimate of the posterior mean field $m=\langle\varphi\rangle_{(\varphi|d,\,\Phi,\,\mathrm{LN})}$
or mean signal $\overline{s}=\langle e^{\varphi}\rangle_{(\varphi|d,\,\Phi,\,\mathrm{LN})}$.
An unknown covariance further complicates this problem considerably. 

\subsection{Likelihood contribution}

The LN-likelihood internal energy and its gradients,

\begin{eqnarray}
U_{\mathrm{LN}}(m,\,D|d) & = & \langle\mathcal{H}(d|\varphi,\,\mathrm{LN})\rangle_{(\varphi|d,\,I)}\nonumber \\
 & \widehat{=} & \mathrm{\frac{1}{2}\,\int dx}\int dy\,\left(R^{\dagger}N^{-1}R\right)_{xy}\nonumber \\
 &  & \times e^{m'_{x}+m'_{y}+D_{xy}}\nonumber \\
 &  & -\left.e^{m'}\right.^{\dagger}R^{\dagger}N^{-1}d,\mbox{ with}\nonumber \\
m' & = & m+\frac{1}{2}\widehat{D},\mbox{ and}\\
\frac{\delta U_{\mathrm{LN}}}{\delta m_{x}} & = & e^{m'_{x}}\left[\int dy\,\left(R^{\dagger}N^{-1}R\right)_{xy}e^{m'_{y}+D_{xy}}\right.\nonumber \\
 &  & -\left(R^{\dagger}N^{-1}d\right)_{x}\biggr],\nonumber \\
\frac{\delta U_{\mathrm{LN}}}{\delta D_{xy}} & = & \frac{\delta_{xy}}{2}\,e^{m'_{x}}\left[\int dz\,\left(R^{\dagger}N^{-1}R\right)_{xz}e^{m_{z}'+D_{xz}}\right.\nonumber \\
 &  & \left.-\left(R^{\dagger}N^{-1}d\right)_{x}\right],\nonumber \\
 &  & +\mathrm{\frac{1}{2}\,}\,\left(R^{\dagger}N^{-1}R\right)_{xy}e^{m'_{x}+m'_{y}+D_{xy}}\nonumber 
\end{eqnarray}
are a bit intricate due to the $e^{D_{xy}}$ terms. These would require,
in principle, an explicit calculation of all these matrix elements,
which is currently completely out of reach for megapixel sized field
inference problems requiring to track $10^{12}$ entries of $D$.
We will therefore replace $D_{xy}\rightarrow\frac{\epsilon}{2}\left(\widehat{D}_{x}+\widehat{D}_{y}\right)$,
with $\epsilon\in[0,1].$ $\epsilon=1$ is a reasonable approximation
if the field correlation volume exceeds largely the footprint of the
point spread function (and the noise covariance is diagonal). This
might be the case for highly resolving imaging instruments. $\epsilon=0$
should be a suitable approximation in case it is exactly the other
way around, as it is the case in interferometry and tomography, where
the instrument data are influenced by extended integration regions.

With this approximation, we define $m''=m'+\frac{\epsilon}{2}\,\widehat{D}=m+\frac{1+\epsilon}{2}\,\widehat{D}$
and $R_{m}=R\,\widehat{e^{m}}$ so that 

\begin{eqnarray}
\frac{\delta U_{\mathrm{LN}}}{\delta m} & = & R_{m''}^{\dagger}N^{-1}R\,e^{m''}-R_{m'}^{\dagger}N^{-1}d\equiv-j,\nonumber \\
\frac{\delta U_{\mathrm{LN}}}{\delta D_{xy}} & = & \frac{1}{2}\,\left(R_{m''}^{\dagger}N^{-1}R_{m''}-\widehat{j}\right).
\end{eqnarray}
During the Gibbs free energy decent of the algorithm, the term $-\widehat{j}$
is omitted in the last equation, to ensure positive definiteness of
$S$.

\section{Poisson Log-normal model\label{sec:Poisson-Log-normal-model}}

\subsection{Field measurement}

Here, the signal $s=e^{\varphi}$ is again log-normal, but the measurement
is Poissonian and therefore subject to signal dependent shot noise.
The signal response $R$ only determines the expected number $\mu=\langle d\rangle_{(d|s,\,\mathrm{PLN})}$
of counts according to 
\begin{equation}
\mu=R\,e^{\varphi}+b,
\end{equation}
where $b$ is the background count rate, here assumed to be known.
An individual datum $d_{i}\in\mathbb{N}_{0}$ follows the Poisson
distribution
\begin{equation}
P(d_{i}|\mu_{i})=\frac{\mu_{i}^{d_{i}}}{d_{i}!}e^{-\mu_{i}}
\end{equation}
independently of the other data. Therefore,
\begin{equation}
\mathcal{H}(d|\varphi,\,\mathrm{PLN})\widehat{=}1^{\dagger}R\,e^{\varphi}-d^{\dagger}\ln\left(R\,e^{\varphi}+b\right),\label{eq:PLN}
\end{equation}
where the logarithm has to be applied component wise. The joint Hamiltonian
in case of know covariance 
\begin{align}
\mathcal{H}(d,\,\varphi|\Phi,\,\mathrm{PLN})= & \mathcal{\,H}(d|\varphi,\,\mathrm{PLN})+\mathcal{H}(\varphi|\Phi)\nonumber \\
\widehat{=} & \,\frac{1}{2}\varphi^{\dagger}\Phi^{-1}\varphi\\
 & +1^{\dagger}R\,e^{\varphi}-d^{\dagger}\ln\left(R\,e^{\varphi}+b\right)\nonumber 
\end{align}
also does not permit simple linear estimate of the posterior mean
field $m=\langle\varphi\rangle_{(\varphi|d,\,\Phi,\,\mathrm{PLN})}$
or mean signal $\overline{s}=\langle e^{\varphi}\rangle_{(\varphi|d,\,\Phi,\,\mathrm{PLN})}$. 

It is interesting to note that in case of no background $b=0$, a
local response $R_{ix}=r_{i}\delta(x-x_{i})$ and a restriction of
the field to the measured locations, $\phi=(\varphi(x_{i}))_{i}$
its Hamiltonian 
\begin{equation}
\mathcal{H}(d,\,\phi|\Phi,\,\mathrm{PLN},b=0,r)\widehat{=}\frac{1}{2}\phi^{\dagger}\Phi^{-1}\phi+\ln|\Phi|-d^{\dagger}\phi+r^{\dagger}e^{\varphi}
\end{equation}
 becomes structurally similar to that of the log-spectrum given by
Eq. \eqref{eq:spectral_prior}.

\subsection{Likelihood contribution}

The PLN-likelihood internal energy is
\begin{eqnarray}
U_{\mathrm{PLN}}(m,\,D|d) & = & \langle\mathcal{H}(d|\varphi,\,\mathrm{LN})\rangle_{(\varphi|d,\,I)}\\
 & \widehat{=} & 1^{\dagger}R\,e^{m'}-d^{\dagger}\left\langle \ln\left(R\,e^{m+\phi}+b\right)\right\rangle _{\mathcal{G}(\phi|D)},\nonumber 
\end{eqnarray}
where
\begin{eqnarray}
\left\langle \ln\left(R\,e^{m+\phi}+b\right)\right\rangle _{i} & = & \ln\left(R_{i}^{\dagger}\,e^{m'}+b_{i}\right)\\
 &  & +\left\langle \ln\underbrace{\left(\frac{R_{i}^{\dagger}\,e^{m+\phi}+b_{i}}{R_{i}^{\dagger}\,e^{m'}+b_{i}}\right)}_{(1+\varepsilon_{i})}\right\rangle \nonumber 
\end{eqnarray}
with

\begin{eqnarray}
\left\langle \ln\left(1+\varepsilon_{i}\right)\right\rangle  & = & 0+\underbrace{\left\langle \varepsilon_{i}\right\rangle }_{=0}-\frac{1}{2}\left\langle \varepsilon_{i}^{2}\right\rangle +\ldots\nonumber \\
 & = & \frac{\int dx\int dy\,e^{m'_{x}+m'_{y}}\left(1-e^{D_{xy}}\right)R_{ix}R_{iy}}{2\,\left(R_{i}^{\dagger}\,e^{m'}+b_{i}\right)^{2}}\nonumber \\
 &  & +\ldots
\end{eqnarray}
as correction term.

The gradients are then
\begin{eqnarray}
\frac{\delta U_{\mathrm{PLN}}}{\delta m_{x}} & \approx & e^{m'_{x}}\,\left[\left(1-\frac{d}{\mu}\right)^{\dagger}R\right]_{x}\nonumber \\
 &  & -\sum_{i}\frac{d_{i}}{\mu_{i}^{2}}e^{m'_{x}}R_{ix}\int dy\,e^{m'_{y}}\left(1-e^{D_{xy}}\right)R_{iy}\nonumber \\
 &  & +\sum_{i}\frac{d_{i}}{\mu_{i}^{3}}R_{ix}e^{m'_{x}}\nonumber \\
 &  & \times\left[\int dz\int dy\,e^{m'_{z}+m'_{y}}\left(1-e^{D_{zy}}\right)R_{iz}R_{iy}\right],\nonumber \\
\mbox{ with }\mu & = & R\,e^{m'}+b,\mbox{ and}\nonumber \\
\frac{\delta U_{\mathrm{PLN}}}{\delta D_{xy}} & \approx & \frac{1}{2}\,\delta_{xy}\,e^{m'_{x}}\,\left[\left(1-\frac{d}{\mu}\right)^{\dagger}R\right]_{x}\nonumber \\
 &  & -\sum_{i}\frac{d_{i}}{2\mu_{i}^{2}}\,\delta_{xy}\,e^{m'_{x}}\,R_{ix}\int dy\,e^{m'_{y}}\left(1-e^{D_{xy}}\right)R_{iy}\nonumber \\
 &  & +\sum_{i}\frac{d_{i}}{2\mu_{i}^{3}}\delta_{xy}\,e^{m'_{x}}\,R_{ix}\nonumber \\
 &  & \times\left[\int dz\int dy\,e^{m'_{z}+m'_{y}}\left(1-e^{D_{zy}}\right)R_{iz}R_{iy}\right]\nonumber \\
 &  & +e^{m'_{x}+m'_{y}}e^{D_{xy}}\sum_{i}\frac{d_{i}}{\mu_{i}^{2}}\,R_{ix}R_{iy}
\end{eqnarray}
Using as for the LN case the approximation $D_{xy}\approx\frac{\epsilon}{2}\left(D_{xx}+D_{yy}\right)$
in exponents, the notation $m''=m'+\frac{\epsilon}{2}\,\widehat{D}=m+\frac{1+\epsilon}{2}\,\widehat{D}$
and $R_{m}=R\,\widehat{e^{m}}$ , we get
\begin{eqnarray}
\frac{\delta U_{\mathrm{PLN}}}{\delta m} & \approx & \left(1-\frac{d}{\mu}\right)^{\dagger}R_{m'}\label{eq:PLNx}\\
 &  & -d^{\dagger}\mu^{-2}\left(R_{m'}R\,e^{m'}-R_{m''}R\,e^{m''}\right)\nonumber \\
 &  & +d^{\dagger}\mu^{-3}R_{m'}\left[\left(R\,e^{m'}\right)^{2}-\left(R\,e^{m''}\right)^{2}\right]\nonumber \\
 & \equiv & j\nonumber \\
\frac{\delta U_{\mathrm{PLN}}}{\delta D_{xy}} & \approx & \frac{1}{2}\,\left(R_{m''}^{\dagger}\widehat{d\,\mu^{-2}}R_{m''}-\widehat{j}\right).
\end{eqnarray}
In the implementation of the PLN reconstruction only the first term
in Eq.\ \eqref{eq:PLNx} is actually used, as the others are small
corrections. During the Gibbs free energy decent of the algorithm,
the term $-\widehat{j}$ is omitted in the last equation, to ensure
positive definiteness of $S$. It is, however used when the minimum
is reached, to calculate the correct covariances. Note that 
\begin{eqnarray}
j & \approx & \left(1-\frac{d}{\mu}\right)^{\dagger}R_{m'}\nonumber \\
 &  & +2\epsilon\,d^{\dagger}\mu^{-2}R_{m'}\widehat{D}\,R_{m'}\nonumber \\
 &  & -2\epsilon\,d^{\dagger}\mu^{-3}R\,e^{m'}R_{m'}\widehat{D}\,R_{m'}+\mathcal{O}(\epsilon^{2})\\
 & = & \left(1-\frac{d}{\mu}\right)^{\dagger}R_{m'}+2\epsilon\,d^{\dagger}\mu^{-3}b\,\left(R_{m'}\widehat{D}\right)\,R_{m'}+\mathcal{O}(\epsilon^{2})\nonumber 
\end{eqnarray}
has only a significant contribution of order $\mathcal{O}(\epsilon)$
if the background $b$ is large.

\section{Operator probing\label{subsec:Auxiliary-fields}}

We propose to treat terms like $\widehat{D}$ as independent, dynamically
evolving quantities. Since we rarely have an explicit representations
of $D$, which explicitly carries matrix elements required for these
quantities of interest, an expensive and noisy stochastic estimation
invoking the implicit representation of $D$ has to be performed.
While doing this, we should take care to retain any already obtained
information on these quantities for the sake of reducing noise and
computational costs. 

As detailed below, each such quantity $A\in\{\widehat{D},\,\widetilde{A}_{\mathrm{c}},\,A_{\Theta}\}$
can be estimated stochastically by calculating the trace of a specific
operator $\mathbb{O}_{A}$ by probing
\begin{equation}
A\equiv\mathrm{Tr}\left[\mathbb{O}_{A}\right]=\langle\xi^{\dagger}\mathbb{O}_{A}\,\xi\rangle_{(\xi)},
\end{equation}
where we will use only a finite number of white noise samples $\xi\hookleftarrow\mathcal{P}(\xi)=\mathcal{G}(\xi,\mathbb{1})$
to estimate the average. 

The diagonal $\widehat{D}$ of the field uncertainty dispersions provides
diagnostically valuable uncertainty quantification, as $\varphi_{x}=m_{x}\pm\widehat{D}_{x}^{\nicefrac{1}{2}}$
in colloquial language. Furthermore, knowing $\widehat{D}$ is also
required by the FrEE dynamics in case of the LN- and PLN-likelihoods
(see Appendices \ref{sec:Log-normal-model} and \ref{sec:Poisson-Log-normal-model}).
We write
\begin{equation}
\widehat{D}_{x}\equiv D_{xx}=\mathrm{Tr}\left[*_{x}D\right]=\langle\xi^{\dagger}*_{x}D\,\xi\rangle_{(\xi)}\equiv\langle\xi_{x}\left(D\,\xi\right)_{x}\rangle_{(\xi)},
\end{equation}
where the $*$ operator vector projects out the position space diagonal
of a matrix $D$ if applied to this with the matrix scalar product
$\mathrm{Tr}\left(*D\right)=\widehat{D}$. Thus we have $\mathbb{O}_{\widehat{D}}=*D$.
Since $\widehat{D}\in[0,\widehat{\Phi_{t'}}]$, we clip at these values.

The harmonic space field uncertainty
\[
\widetilde{D}_{k}\equiv\mathrm{Tr}\left[\,\mathbb{P}_{k}D\right]=\langle\xi^{\dagger}\mathbb{P}_{k}\,D\,\xi\rangle_{(\xi)}
\]
can be obtained by stochastic probing the operator $\mathbb{O}_{\widetilde{D}}=\mathbb{P}\,D$.
While doing so, we should use a minimum of prior information on the
values of $\widetilde{D}$. First $\widetilde{D}\ge0$ since $D$
is a covariance. From $D\le\Phi=r\,\mathbb{P}^{\dagger}e^{t'}$ we
conclude 
\begin{equation}
\widetilde{D}\le\mathrm{Tr}\left[\mathbb{P}\,\Phi\right]=r\,\varrho\,e^{t'}\equiv\widetilde{\Phi}_{t'}
\end{equation}
and therefore enforce $\widetilde{D}_{k}\in[0,\,\widetilde{\Phi}_{k}]$
by clipping any value outside of this interval.

\bibliographystyle{apsrev4-1}
\bibliography{ift}

\end{document}